%% file: main.tex
\documentclass[twoside,11pt]{article}

\usepackage{jmlr2e}

\usepackage{multirow}

\usepackage{enumitem}
\usepackage{bm}%
\usepackage{amsmath}%
\usepackage{amsfonts}%
\usepackage{amssymb}%
\usepackage{colonequals}
\usepackage{psfrag}
\usepackage{mathrsfs}
\usepackage{xcolor}
\usepackage{hyperref}

\usepackage{booktabs,siunitx}

\usepackage{algorithm}
\usepackage{algpseudocode}
\usepackage{comment}

\hypersetup{%
  colorlinks=true,
  urlcolor=blue,%
  linkcolor=blue,%
  citecolor=blue,%
  linkbordercolor=blue,
  pdfborderstyle={/S/U/W 1}
}

\input{notation.tex}

\usepackage{lastpage}
\jmlrheading{}{2024}{1-\pageref{LastPage}}{null}{null}{}{Alim$^*$, Wang$^*$, Gulati, Srivastava, and Azizan}

\ShortHeadings{Differentially Private Synthetic Data Generation for Relational Databases}{Alim$^*$, Wang$^*$, Gulati, Srivastava, and Azizan}

\firstpageno{1}

\begin{document}

\title{Differentially Private Synthetic Data Generation for Relational Databases} 

\author{\name Kaveh Alim\thanks{Equal contribution.} \email mrz@mit.edu \\
       \addr MIT
       \AND
       \name Hao Wang\footnotemark[1] \email hao-wang@redhat.com \\
       \addr RedHat AI Innovation \& MIT-IBM Watson AI Lab
       \AND
       Ojas Gulati  \email ojasg@mit.edu\\
       \addr MIT
       \AND
       \name Akash Srivastava \email akash@redhat.com \\
       \addr RedHat AI Innovation \& MIT-IBM Watson AI Lab
       \AND
       Navid Azizan \email azizan@mit.edu \\
       \addr MIT 
       }

\editor{}

\maketitle

\begin{abstract}%
Existing differentially private (DP) synthetic data generation mechanisms typically assume a single-source table. In practice, data is often distributed across multiple tables with relationships across tables. In this paper, we introduce the first-of-its-kind algorithm that can be combined with \emph{any} existing DP mechanisms to generate synthetic relational databases. Our algorithm iteratively refines the relationship between individual synthetic tables to minimize their approximation errors in terms of low-order marginal distributions while maintaining referential integrity. This algorithm eliminates the need to flatten a relational database into a master table (saving space), operates efficiently (saving time), and scales effectively to high-dimensional data. We provide both DP and theoretical utility guarantees for our algorithm. Through numerical experiments on real-world datasets, we demonstrate the effectiveness of our method in preserving fidelity to the original data.
\end{abstract}

\input{section.tex}

\clearpage
\appendix
\input{appendix.tex}

\clearpage
\subsubsection*{Acknowledgments}
This work was supported in part by the MIT-IBM Watson AI Lab. The authors acknowledge the MIT SuperCloud and Lincoln Laboratory Supercomputing Center for providing computing resources that have contributed to the results reported within this paper.
\bibliography{main}

\end{document}

%% file: notation.tex
\DeclareMathOperator*{\argmin}{\arg\!\min}

\newcommand{\ba}{\bm{a}}
\newcommand{\bb}{\bm{b}}

\newcommand{\bp}{\bm{p}}
\newcommand{\bq}{\bm{q}}

\newcommand{\bx}{\bm{x}}

\newcommand{\by}{\bm{y}}
\newcommand{\bv}{\bm{v}}

\newcommand{\bw}{\bm{w}}

\newcommand{\bI}{\bm{I}}
\newcommand{\bM}{\bm{M}}

\newcommand{\bB}{\bm{B}}

\newcommand{\bQ}{\bm{Q}}

\newcommand{\ones}{\bm{1}}

\renewcommand{\tilde}{\widetilde}

\newcommand{\Reals}{\mathbb{R}}
\newcommand{\defined}{\triangleq}

\newcommand{\sto}{\mbox{\normalfont s.t.}}

\newcommand{\indicator}[1]{\mathbb{I}({#1})}

\newcommand{\ExpVal}[2]{\mathbb{E}\left[ #2 \right]}
\newcommand{\EE}[1]{\ExpVal{}{#1}}

\newcommand{\mse}{\mathsf{mse}}

\newcommand{\Var}[1]{\mathsf{Var}\left(#1\right)}

\newcommand{\floor}[1]{\lfloor #1 \rfloor}

\newcommand{\groupsum}[2][\mathcal{G}]{\mathsf{sum}(\bx\mid #1[#2])}

%% file: section.tex
\section{Introduction}

Relational databases play a pivotal role in modern information systems and business operations due to their efficiency in managing structured data \citep{schleich2019learning}. 
According to a Kaggle survey \citep{kaggle17}, 65.5\% of users worked extensively with relational data. Additionally, the majority of leading database management systems (e.g., MySQL and Oracle) are built on relational database principles \citep{DBengines}. 
These systems organize data into multiple tables, each representing a specific entity, and the relationships between tables delineate the connections between these entities. 
However, the widespread use of relational databases also carries a significant risk of privacy leakage. 
For example, if a single table suffers from a privacy breach, all other tables containing sensitive information can be exposed, as they are interconnected and share relationships. Moreover, deleting data in a relational database can be complex, and incomplete deletion can leave traces of data, which can potentially be accessed and reconstructed by attackers.

Today, differential privacy (DP) stands as the de facto standard for privacy protection \citep{abowd2018us,whitehouse23}. There is a growing body of research focused on applying DP to generate private synthetic data \citep[see, e.g.,][]{ridgeway2021challenge,mckenna2021winning,liu2021iterative,opendp2023}. This effort enables data curators to share (synthetic) data while ensuring the privacy of individuals' personal information within the original dataset.  In return, they can harness advanced machine learning (ML) techniques employed by end-users trained on the synthetic data. Numerous studies have provided evidence that state-of-the-art (SOTA) DP synthetic data effectively preserves both the statistical properties of the original data and the high performance of downstream predictive models trained on these synthetic datasets when deployed on the original data \citep{tao2021benchmarking,wang2023postprocessing}. 
However, all existing works assume a single-source database (with a few exceptions discussed in Related Work). 
This motivates the central question we aim to tackle in this paper:
\begin{center}
\emph{Can we adapt existing DP synthetic data generation algorithms to relational databases while preserving their referential integrity?}
\end{center}

The conventional approach---flattening relational databases into a single master table, generating a synthetic master table, and dividing it into separate databases---presents several challenges. 
To illustrate, consider education data as an example, with two tables (student and teacher information) linked by students' enrollment in a teacher's course. 
First, flattening may lead to data integrity issues by introducing numerous null values. This makes it difficult to distinguish between missing and intentionally null data. For instance, if a teacher is not currently teaching, their students' information in the master table will be represented by null values. Consequently, the master synthetic table will contain a significant number of null values even if the original tables do not have any null values.
Second, flattening introduces concerns related to data scalability and redundancy. Each record in the master table may have an extremely large number of features, which poses challenges for SOTA DP synthetic data generation mechanisms, as their running time grows rapidly as the number of features increases \citep[e.g.,][]{mckenna2022aim,mckenna2021winning,vietri2022private,aydore2021differentially}. Additionally, certain records from the original tables might be duplicated across multiple entries in the master table, resulting in data redundancy and an increased demand for storage space. 
Third, breaking down a synthetic master table may disrupt relationships, especially when there are records with shared feature values. For example, if the master table includes two rows with identical student demographic information, it is unclear whether they are the same student or different students with matching demographic details. 
Finally, since individual records may repeat multiple times in the master table, the sensitivity of statistical queries can significantly increase, potentially becoming unbounded. Consequently, applying existing DP mechanisms may result in a synthetic master table with limited utility.

The goal of this paper is to introduce the first-of-its-kind algorithm that can generate synthetic relational databases while preserving the privacy, statistical properties, and referential integrity of the original data. For this purpose, we first extend the definition of $k$-way marginal queries \citep{thaler2012faster,ridgeway2021challenge} to relational databases and articulate the requirements of referential integrity (Section~\ref{sec::prel}). In the context of relational databases, $k$-way marginal queries measure the low-order marginal distributions of features (i.e., columns) across different tables. Preserving these distributions is essential for many downstream applications of synthetic data, such as data visualization, building data pre-processing pipelines, and training ML predictive models like logistic regression.

Our main contribution is an algorithm that effectively generates synthetic relational data by minimizing approximation errors in terms of $k$-way marginal queries (Algorithm~\ref{alg::adapt_alg_l2}). 
This algorithm works as follows. 
First, we apply an off-the-shelf DP mechanism to generate individual synthetic tables. Then we represent the relational synthetic database as a bipartite graph and establish relationships by learning its bi-adjacency matrix. We prove that $k$-way marginals can be reformulated as (fractional) linear functions of this bi-adjacency matrix (Lemma~\ref{lem::qv}). Based on this observation, we propose a combinatorial optimization that minimizes the worst-case approximation error of marginal queries between the original and synthetic databases (Section~\ref{sec::main}). The combinatorial optimization is solved using an iterative algorithm. At each step, we identify a subset of $k$-way marginal queries with the highest approximation errors. Then we solve a relaxed optimization in continuous space to update the bi-adjacency matrix (Section~\ref{sec::proj_grad_desc}). Following this, we introduce an unbiased sampling algorithm to convert the learned bi-adjacency matrix back into discrete space (Section~\ref{sec::ubs}). To generate large-scale synthetic data, we propose a random slicing heuristic that leverages the sparsity of the bi-adjacency matrix, allowing us to optimize bi-adjacency matrices with millions of parameters (Section~\ref{apx::scalability}).

Our approach offers several benefits. 
First, it is highly flexible, as it can be combined with \emph{any} off-the-shelf DP synthetic data generation mechanisms. These DP mechanisms are used to generate individual synthetic tables, and our algorithm can then be applied to build relationships between these tables.
Second, it avoids the need to flatten relational databases into a master table; instead, it only requires querying the relational databases to compute low-order marginal distributions, which can be done using SQL aggregate functions. 
Third, we establish convergence guarantees for our algorithm and prove that our algorithm has linear runtime complexity with respect to the size of the synthetic data (Theorem~\ref{thm::convergence} and \ref{thm::ubs}). These properties, combined with the random slicing heuristic, enable running our algorithm to generate large-scale synthetic data. 
Finally, we present utility theorems that upper bound the approximation error for estimating $k$-way marginals using synthetic data generated by our method (Theorem~\ref{thm::concentration} and~\ref{thm::Projection_utility}). The proofs leverage properties of negatively associated random variables \citep{Joagdev1983NegativeAO}  to establish Hoeffding's inequality for dependent variables, along with classical tools from convex geometry and concentration inequalities. 
We believe these new proof techniques for proving properties of synthetic data could be of independent interest to the DP synthetic data community.

We conduct numerical experiments to validate our method. 
We couple the iterative algorithm with SOTA DP synthetic data generation mechanisms to generate relational synthetic databases. We present a scalable PyTorch implementation of our algorithm suited for high-dimensional data. The results show that our algorithm effectively maintains key statistical properties of the original relational database to a significant extent. 
Note that existing literature on synthetic relational data generation, even without DP guarantees, is limited \citep{patki2016synthetic,mami2022generating,yang2022sam,xu2023synthetic,cai2023privlava,francis2024comparison,pang2024clavaddpm}. We hope our efforts (e.g., introducing new benchmark datasets and open-sourcing PyTorch code) 
can benefit the communities focused on DP synthetic data and inspire new research on this topic.

\subsection*{Related Work}

Privacy-preserving synthetic tabular data generation is an active research topic \citep[see e.g.,][]{blum2013learning,ullman2020pcps,boedihardjo2022privacy,chen2015differentially,ge2020kamino,rosenblatt2020differentially,wang2023postprocessing,he2023algorithmically,donhauser2024privacy,greenewald2024privacy,bun2024continual}. For example, a line of work consider learning probabilistic graphical models \citep{zhang2017privbayes,mckenna2019graphical,mckenna2021winning} or generative adversarial networks \citep{xie2018differentially,beaulieu2019privacy,jordon2019pate,tantipongpipat2019differentially,neunhoeffer2020private,bie2023private} with DP guarantees and generating synthetic data by sampling from these models. Another line of work propose to iteratively refine synthetic data to minimize its approximation error on a pre-selected set of workload queries \citep{hardt2012simple,gaboardi2014dual,mckenna2022aim,liu2021iterative,liu2021leveraging,vietri2020new,aydore2021differentially,vietri2022private,liu2023generating}.

Most existing work on DP synthetic data generation only focuses on a single-source dataset without considering relational database. The only exceptions are \citet{xu2023synthetic,cai2023privlava}. Among them, \citet{cai2023privlava} uses graphical models for generating DP synthetic data, incorporating latent variables to capture the primal-foreign-key relationship\footnote{A primary key is a column in the parent table that uniquely identifies each row. A foreign key is a column in the child table that creates relationships between the two tables by referencing the primary key in the parent table.}. However, their approach is restricted to one-to-many relationships (i.e., records in one table have exactly one relationship with the other table) and can only produce synthetic relational data of this type. As discussed in Appendix~\ref{apx::one-to-many}, with only one-to-many relationships, the structure of relational data is simpler, making the synthetic data generation process more straightforward and computationally efficient. In contrast, our method can accommodate any type of relationship, including the more complex many-to-many cases (i.e., records in one table can be related to multiple records in another table, and vice versa). 
A more closely related work is \citet{xu2023synthetic}, which can produce multiple tables with many-to-many relationships by leveraging tools from random graph theory and representation learning. However, their method requires using DP-SGD to optimize their generative models for each table generation, whereas our method can seamlessly integrate with \emph{any} existing DP mechanisms for synthetic table generation. This versatility is essential, particularly as marginal-based and workload-based mechanisms often produce higher-quality synthetic tabular data than DP-SGD-based mechanisms \citep{tao2021benchmarking,mckenna2021winning,wang2023postprocessing}. Finally, some research explores DP in relational database systems, primarily focusing on releasing statistical queries \citep[see e.g.,][]{he2021complex,johnson2018towards,kotsogiannis2019privatesql}. In contrast, our emphasis is on releasing a synthetic copy of the relational database, extending the scope of privacy-preserving data generation methods to a more practical and general setting.

Another line of work focused on the release of DP synthetic graphs \citep{liu2023optimal,zhang2021differentially,eliavs2020differentially,upadhyay2013random,gupta2012iterative,blocki2012johnson,sala2011sharing,hay2009accurate,liu2024pgb} with the goal of maintaining essential graph properties (e.g., connectivity, degree distribution, and cut approximation). While a relational database can often be represented as a multipartite graph (see Section~\ref{subsec::bipar_graph}), applying these existing approaches to generate synthetic relational databases encounters various challenges. First, most existing methods assume that the vertex set does not have any attribute information and is the same between real and synthetic graphs. They only focus on learning the edge set through DP mechanisms. In our case, however, the vertex sets (representing records) include attribute information (feature values of records) and differ between real and synthetic databases, requiring privacy preservation for both the vertex attributes and the edge set. 
The only exceptions are \citet{chen2019publishing,jorgensen2016publishing}, which consider vertex attributes when generating DP synthetic graphs. 
However, \citet{chen2019publishing} only produces synthetic graphs with the same number of vertices as the original, while applying \citet{jorgensen2016publishing} to generate synthetic relational databases can lead to referential integrity violation (e.g., one-to-many relationships in the original data could become many-to-many in the synthetic data). 
Finally, existing approaches prioritize preserving graph properties, whereas our objective extends to preserve both graph properties (i.e., relationships between different tables) and statistical properties of relational databases (i.e., low-order marginal distributions).

Broadly, a body of work has investigated applications of relational databases without imposing privacy constraints \citep{patki2016synthetic,mami2022generating,yang2022sam,francis2024comparison,pang2024clavaddpm}. For instance, some works focus on training predictive models on relational data \citep{cvitkovic2020supervised,gan2024graph}, while others aim to analyze and extract representations from these datasets \citep{cai2021arm,fey2023relational}. There is also research on generating synthetic relational databases, but these efforts are typically restricted to handling one-to-many relationships defined by primary-foreign keys \citep{patki2016synthetic,mami2022generating,pang2024clavaddpm}. Additionally, some studies have introduced benchmarks for evaluating deep learning models on relational databases \citep{robinson2024relbench}. In contrast, our work focuses on synthetic relational data generation that supports both one-to-many and many-to-many relationships. Furthermore, we incorporate DP guarantees into the data generation process to ensure robust privacy protection for the original relational databases.

\section{Preliminaries and Problem Formulation}
\label{sec::prel}

We introduce notation, represent relational databases through bipartite graphs, review differential privacy and its properties, and provide an overview of our problem formulation.

\subsection{Bipartite Graph and Relational Database}
\label{subsec::bipar_graph}

To simplify our presentation, we assume that the relational database $\mathcal{B}$ consists of only two tables $\mathcal{D}_1$ and $\mathcal{D}_2$ (e.g., student and teacher information), each having $n_1$ and $n_2$ rows. 
We represent the relational database as a bipartite graph $\mathcal{B} = (\mathcal{D}_1,\mathcal{D}_2,\bB)$ where each record corresponds to a node; the two tables define two distinct sets of nodes; and the relationships between the tables are depicted as edges connecting these nodes. We represent the edges using a bi-adjacency matrix, denoted as $\bB \in \{0,1\}^{n_1 \times n_2}$, where $B_{i,j} = 1$ iff the $i$-th record from table $\mathcal{D}_1$ is connected to the $j$-th record from table $\mathcal{D}_2$ (e.g. if the $i$-th student is enrolled in the $j$-th teacher's course). For each record $\bx$, we represent its degree as the total number of records from the other table connected to $\bx$, denoted by $\deg(\bx)$. Finally, we assume the degree of each record is upper bounded by a constant $d_{\max}$.

\subsection{Differential Privacy}
\label{sec::dp}

We recall the definition of differential privacy (DP) \citep{dwork2014algorithmic} and provide detailed background information in Appendix~\ref{apx::dp}.
\begin{definition}
\label{def::reldp}
A randomized mechanism $\mathcal{M}$ that takes a relational database as input and returns an output from a set $\mathcal{R}$ satisfies $(\epsilon, \delta)$-differential privacy, if for any adjacent relational databases $\mathcal{B}$ and $\mathcal{B}'$ and all possible subsets $\mathcal{O}$ of $\mathcal{R}$, we have
\begin{align}
    \mathbb{P}(\mathcal{M}(\mathcal{B}) \in \mathcal{O})
    \leq e^{\epsilon} \mathbb{P}(\mathcal{M}(\mathcal{B}') \in \mathcal{O}) + \delta.
\end{align}
\end{definition}
We consider two relational databases $\mathcal{B}$ and $\mathcal{B}'$ adjacent if $\mathcal{B}'$ can be obtained from $\mathcal{B}$ by selecting a table, modifying the values of a single row within this table, and changing all relationships associated with this row.

DP plays a pivotal role in answering statistical queries, with two distinct categories to be considered for relational databases. The first class involves queries pertaining to a single table, say table $\mathcal{D}_1$. In the context of a student-teacher database, an example would be a query about whether a student is a freshman. If we denote the data domain of table $\mathcal{D}_1$ by $\mathcal{X}_1$, a statistical query can be represented by a function $q: \mathcal{X}_1\to \{0,1\}$ and its average over a database is denoted as $q(\mathcal{B}) \defined \frac{1}{n_1} \sum_{\bx \in \mathcal{D}_1} q(\bx)$. The second class includes cross-table queries, such as whether a teacher and a student belong to the same department. Analogously, these queries are represented by a function $q: \mathcal{X}_1 \times \mathcal{X}_2 \to \{0,1\}$. We denote the average of their values over a database by $q(\mathcal{B}) \defined \frac{1}{\ones^T \cdot \operatorname{vec}(\bB)}\sum q(\bx_i,\bx_j)$ where the sum is over $(\bx_i, \bx_j) \in \mathcal{D}_1\times\mathcal{D}_2$ s.t. $B_{i,j}=1$ and $\operatorname{vec}(\cdot)$ converts a matrix into a vector. When $(\mathcal{D}_1, \mathcal{D}_2)$ are clear from the context, we also express the query in terms of the bi-adjacency matrix $\bB$. 

\subsection{Problem Formulation}
\label{subsec::setup}

Our goal is to generate a privacy-preserving synthetic database $\mathcal{B}_{\text{syn}} = (\mathcal{D}_1^{\text{syn}}, \mathcal{D}_2^{\text{syn}}, \bB^{\text{syn}})$ that preserves both statistical properties and referential integrity of the original relational database. In terms of statistical properties, our aim is to ensure that the marginal distributions of subsets of features in the synthetic data closely align with those of the real data. To achieve this, we revisit the definition of $k$-way marginal query and extend it to suit relational databases. For any integer $d\geq 1$, we define $[d] \defined \{1,\cdots,d\}$.
\begin{definition}
\label{defn::k-way_marginal}
Suppose the domain of table $\mathcal{D}_i$ has $d_i$ categorical features: $\mathcal{X}_i = \mathcal{X}_{i,1} \times \cdots \mathcal{X}_{i,d_i}$. We define a (single-table) $k$-way workload as a subset of $k$ features: $\mathcal{S} \subseteq [d_1]$ (or $\mathcal{S} \subseteq [d_2]$) with $|\mathcal{S}| = k$. Additionally, given a reference value $\by$, we define a (single-table) $k$-way marginal query as
\begin{align}
    q_{\mathcal{S}, \by}(\bx) \defined \prod_{j \in \mathcal{S}} \indicator{x_j = y_j} \quad \text{for }\bx \in \mathcal{X}_1 \ (\text{or } \mathcal{X}_2)
\end{align}
where $\mathbb{I}$ is an indicator function. Analogously, we define a (cross-table) $k$-way workload as $\mathcal{S}_1 \times \mathcal{S}_2 \subseteq [d_1] \times [d_2]$ with $|\mathcal{S}_1| + |\mathcal{S}_2| = k$, $0 < |\mathcal{S}_1| < k$; additionally, given a reference value $(\by_1,\by_2)$, we define a (cross-table) $k$-way marginal query as
\begin{align*}
    q_{(\mathcal{S}_1, \mathcal{S}_2), (\by_1, \by_2)}(\bx_1, \bx_2) \defined \prod_{(j_1,j_2) \in \mathcal{S}} \indicator{(x_{1,j_1}, x_{2,j_2}) = (y_{1,j_1}, y_{2,j_2})} \quad \text{for }(\bx_1,\bx_2) \in \mathcal{X}_1 \times \mathcal{X}_2.
\end{align*}
We denote the set of all (cross-table) $k$-way marginal workloads by $\mathcal{W}_{\text{cross},k}$ and their associated queries by $\mathcal{Q}_{\text{cross},k}$. 
\end{definition}
Our main objective in generating synthetic relational databases is to preserve $k$-way marginal queries. These low-order statistics are useful for many downstream applications of synthetic data. For example, data visualization (e.g., histograms of each column and correlation matrices), data pre-processing pipelines (e.g., normalizing columns using their mean and standard deviation), and training simple ML models (e.g., linear models) all depend on low-order statistics. Therefore, preserving $k$-way marginals in synthetic data allows downstream users to maintain statistical insights extracted from synthetic data and train ML models that perform comparably (within a tolerance) to those trained on real data.
Another example highlighting the importance of $k$-way marginals comes from the National Institute of Standards and Technology (NIST), which organized a competition in 2018 to emphasize the need for privacy preservation in synthetic data generation \citep{mckenna2021winning,ridgeway2021challenge}. In this competition, 3-way marginals were used as a utility metric for evaluating the quality of synthetic data. Here we extend the definition of $k$-way marginals to multiple synthetic tables with relationships.

In maintaining referential integrity, we use a linking table to establish relationships. It is designed to store the IDs of synthetic tables: the entry $(i, j)$ is included in this table if $B_{i,j}^{\text{syn}} = 1$. Moreover, when the original database exhibits a one-to-many relationship (i.e., a record in the parent table can be related to one or more records in the child table, but a record in the child table can be related to \emph{only one} record in the parent table) or one-to-one relationship\footnote{We assume that information about the types of relationships in the original database (e.g., one-to-many or many-to-many) is publicly available.}, we expect the synthetic database to preserve this relationship. Finally, in the case of a one-to-many relationship in the original database, we ensure there are no orphaned rows in the synthetic database. Specifically, we impose the requirement that each record in the child table connects to a record in the parent table (although records in the parent table are permitted to have no associated child records).

\section{Main Results}
\label{sec::main}

\input{algorithm.tex}

\section{Numerical Experiments}
\label{sec::experiments}

We demonstrate our algorithm through numerical experiments on real-world datasets. The code for reproducing our results is available at \url{https://github.com/azizanlab/dp-relational}

\paragraph{Data.} We consider the \texttt{MovieLens} dataset \citep{harper2015movielens} and the \texttt{IPUMS} dataset \citep{ruggles2021ipums}. The \texttt{MovieLens} dataset consists of three tables: users, movies, and ratings. The user table contains 6,040 entries, detailing demographic information like gender, age, and occupation. The movie table contains 3,883 entries, each tagged with associated genres. We convert the genres into 18 binary features---this is not one-hot encoding, as each movie can belong to multiple genres. The ratings table links users and movies in a many-to-many structure: users can rate multiple movies, and each movie can have multiple ratings.  We pre-process the rating table to ensure that each movie is rated by no more than 10 users, and each user rates no more than 10 movies (i.e., $d_{\max} = 10$). After this pre-processing, the rating table contains 10,075 rows. 

The \texttt{IPUMS} dataset contains anonymized U.S. census data. For our experiment, we focus on the user's table, which provides detailed individual-level data, including demographic, educational, and employment attributes containing 3,373,378 entries. To expedite the training, we only consider the first 5\% (168,669 rows) of the records. To establish a many-to-many relationship within this dataset, we use parental links, as the dataset includes the person IDs of each individual's parents. An individual in the \texttt{IPUMS} dataset may have up to four recorded parents—two mothers and two fathers—leading us to set $d_{\max} = 4$. This design creates a self-referential structure within the users table, as the relational connections are entirely within the same dataset table. For the \texttt{IPUMS} dataset, we generate a synthetic users table of 10,000 records.

\paragraph{Setup.} We use state-of-the-art DP mechanisms---\texttt{AIM} \citep{mckenna2022aim}, \texttt{MST} \citep{mckenna2021winning}---to generate individual synthetic tables. Each synthetic table is created with a privacy budget of $\epsilon = 1$. We use the OpenDP library \citep{opendp2023} to implement these DP mechanisms. 
To build relationships within the synthetic tables, we apply Algorithm~\ref{alg::adapt_alg_l2} under varying privacy budgets. We evaluate its performance by measuring the average error in $3$-way cross-table marginal queries (Definition~\ref{defn::k-way_marginal}) between synthetic and real data.

In our ablation study, we vary several parameters. By default, our settings are as follows: the algorithm iterates 15 times, using randomized slicing to create sub-tables of 1,000 synthetic records by 1,000 synthetic records. During each iteration, 3 slices are generated for IPUMS, while 8 are generated in MovieLens, with each slice containing at least 200 rows and 200 columns (20\% of the slice) that include relational data to ensure a sufficient number of relationships in each subset. Within each iteration, the exponential mechanism, with a weight of $\alpha = 0.2$, selects 3 workloads for evaluation. From the queries selected by the exponential mechanism, we choose 8 of them with the highest error for updating each slice using projected gradient descent.

\paragraph{Results.} We demonstrate the performance of our algorithm across various datasets and DP mechanisms in the figures below. Additionally, we conduct an ablation study to assess the impact of different hyper-parameter selections on our results.
\begin{figure}[ht]
    \centering
    \includegraphics[width=0.45\linewidth]{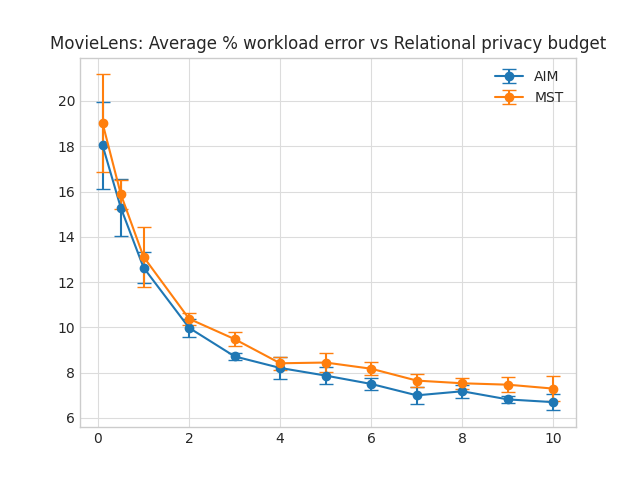}
    \includegraphics[width=0.45\linewidth]{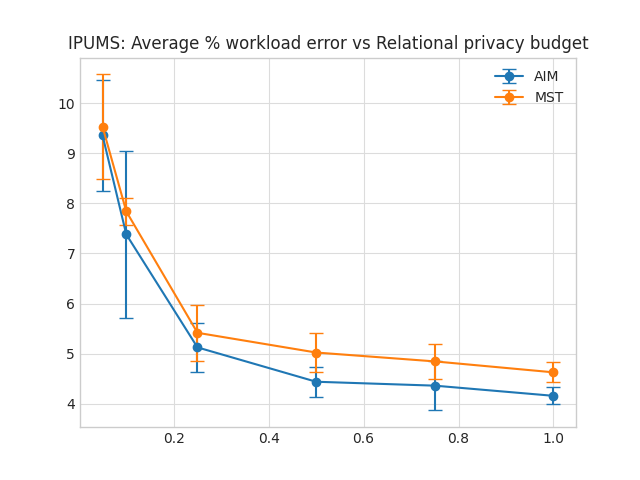}
    \caption{We use \texttt{AIM} and \texttt{MST} to produce individual synthetic tables and then apply our algorithm with privacy budget $\epsilon_{\text{rel}}$ to establish their relationships. We illustrate the impact of varying $\epsilon_{\text{rel}}$ on the average $3$-way marginal query error for the \texttt{MovieLens} dataset (Left) and the \texttt{IPUMS} dataset (Right). As expected, the average error decreases as $\epsilon_{\text{rel}}$ increases, due to reduced noise.}
    \label{fig:eps-ipums}
\end{figure}

\begin{itemize}[leftmargin=1em]
\item \textbf{Privacy budget.} Recall that $\epsilon_{\text{rel}}$ is the privacy budget allocated to Algorithm~\ref{alg::adapt_alg_l2} for establishing relationships among synthetic tables. In Figure~\ref{fig:eps-ipums}, we illustrate the effect of varying $\epsilon_{\text{rel}}$ on the average $3$-way marginal query error. As $\epsilon_{\text{rel}}$ increases, the average error decreases due to reduced noise in both the selection (exponential mechanism) and observation (Gaussian mechanism) phases. This trend is clearly observed in results for both \texttt{AIM} and \texttt{MST}. Furthermore, higher $\epsilon_{\text{rel}}$ values lead to a reduction not only in the average error but also in its standard deviation, resulting from the lower noise levels. Finally, with a small privacy budget (e.g., $\epsilon=1$ for generating each synthetic table and $\epsilon_{\text{rel}} = 2.0$ for establishing relationships), our algorithm is capable of producing a synthetic relational database with high fidelity.
\end{itemize}

\begin{figure}[ht]
    \centering
    \includegraphics[width=0.45\linewidth]{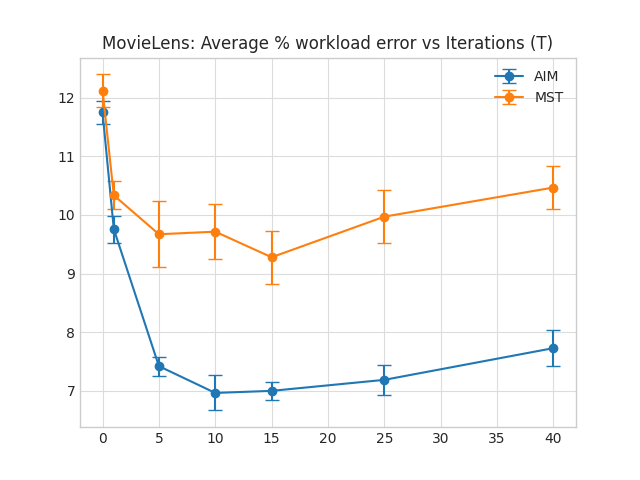}
    \includegraphics[width=0.45\linewidth]{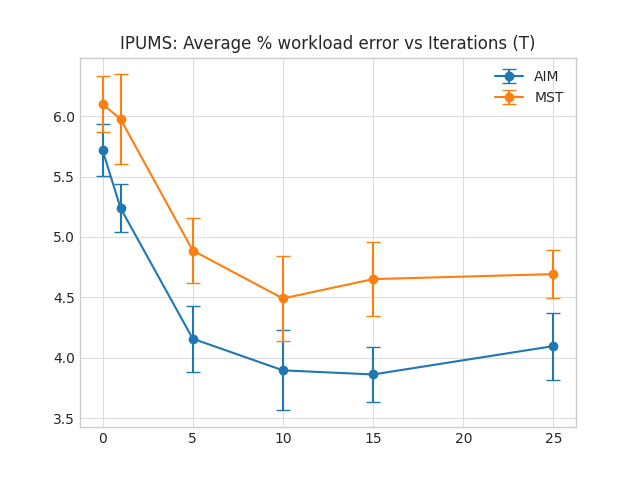}
    \caption{We analyze the impact of varying the number of iterations $T$ in Algorithm~\ref{alg::adapt_alg_l2} on the quality of synthetic data. The figures show a U-shaped curve between average error and $T$. This trend highlights a trade-off: increasing $T$ allows for more workload queries being selected but their answers, computed from real data, become increasingly noisy.}
    \label{fig:T-ipums}
\end{figure}

\begin{itemize}[leftmargin=1em]
\item \textbf{Number of iterations.} 
We analyze how varying the number of iterations $T$ in Algorithm~\ref{alg::adapt_alg_l2} influences synthetic data quality, specifically in terms of the average 3-way marginal query error. Figure~\ref{fig:T-ipums} shows a U-shaped curve between average error and $T$. This effect arises because the privacy budget $\epsilon_{\text{rel}}$ is evenly distributed across $T$ iterations used by the Gaussian and exponential mechanisms. Hence, increasing $T$ allows more workloads to be selected and optimized for synthetic data with the cost of adding higher noise to the workload query answers computed from the real data. Conversely, with very few iterations, the algorithm has limited chances to explore sufficient workload queries, leading to higher average error. This trade-off underpins the U-shaped curve, identifying an optimal range between $T = 10$ and $T=15$ where the error is minimized.  
\end{itemize}

\begin{figure}[ht]
    \centering
    \includegraphics[width=0.45\linewidth]{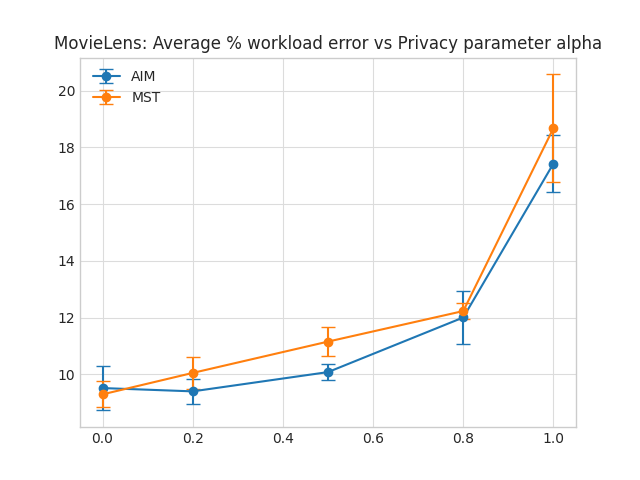}
    \includegraphics[width=0.45\linewidth]{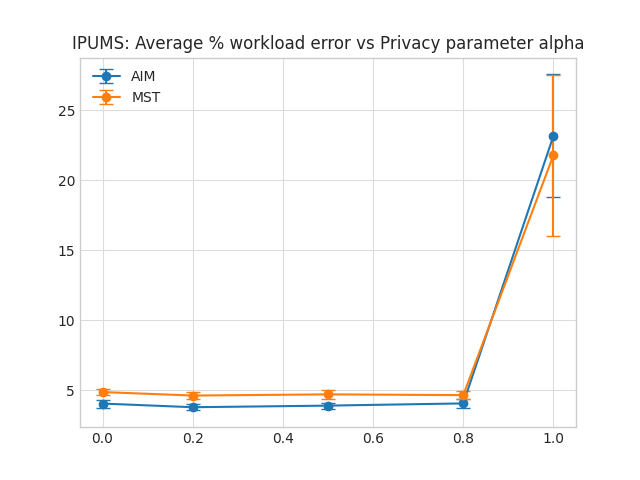}
    \caption{We show the effect of the hyperparameter $\alpha \in [0,1]$ on the quality of synthetic data produced by our algorithm. This parameter allocates the privacy budget between the Gaussian and exponential mechanisms.}
    \label{fig:alpha-ipums}
\end{figure}

\begin{itemize}[leftmargin=1em]
\item \textbf{Privacy budget allocation.} 
In Algorithm~\ref{alg::alg_sparse_large}, a detailed version of our algorithm, we introduce a hyperparameter, $\alpha \in [0,1]$, to divide the privacy budget between the Gaussian and exponential mechanisms. When $\alpha = 1.0$, the entire privacy budget is allocated to the exponential mechanism, enabling it to select the most relevant workloads. However, with no privacy budget assigned to the Gaussian mechanism, it outputs random values for the selected workload queries. Conversely, setting $\alpha = 0.0$ dedicates the entire privacy budget to the Gaussian mechanism, producing low-noise query values derived from real data for randomly selected workloads. Figure~\ref{fig:alpha-ipums} illustrates the impact of varying $\alpha$ on the quality of the synthetic data. Based on this result, we select $\alpha = 0.2$ for the rest of our ablation study.
\end{itemize}

\begin{figure}[ht]
    \centering
    \includegraphics[width=0.45\linewidth]{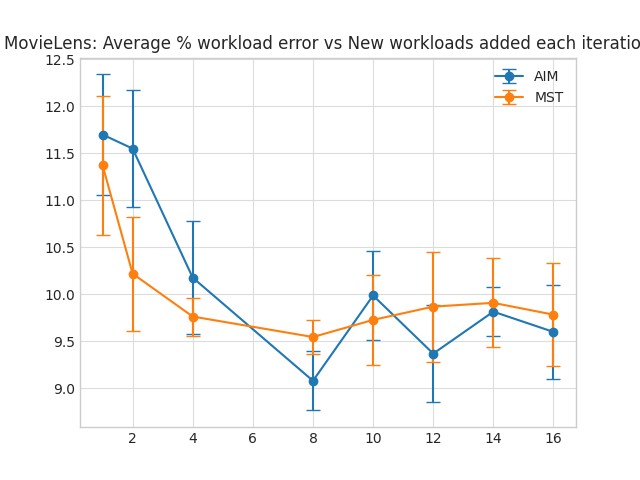}
    \includegraphics[width=0.45\linewidth]{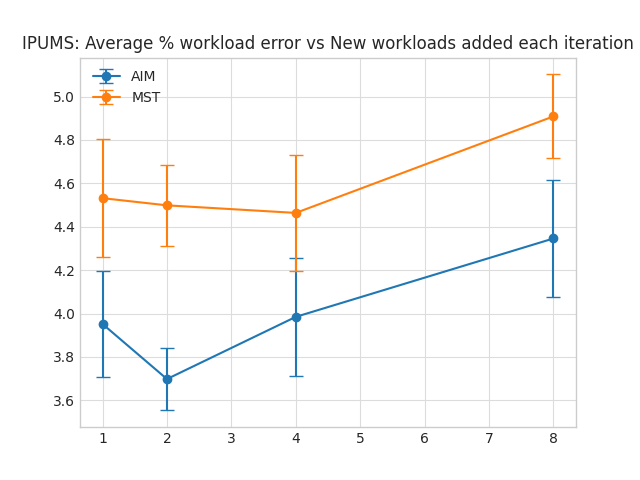}
    \caption{We demonstrate the effect of the number of workloads $K$ selected by the exponential mechanism on the quality of synthetic data produced by our algorithm.}
    \label{fig:nqexp-ipums}
\end{figure}

\begin{itemize}[leftmargin=1em]
\item \textbf{Number of workloads selected by the exponential mechanism.} 
The exponential mechanism in Algorithm~\ref{alg::adapt_alg_l2} selects $K$ workloads per iteration. With the privacy budget distributed evenly across these selected workloads, a larger $K$ increases the noise added to their query answers computed from real data, while a smaller $K$ may yield an insufficiently diverse set of observed workloads. Figure~\ref{fig:nqexp-ipums} illustrates the impact of different values of $K$ on the average $3$-way marginal error.
\end{itemize}

\begin{figure}[ht]
    \centering
    \includegraphics[width=0.45\linewidth]{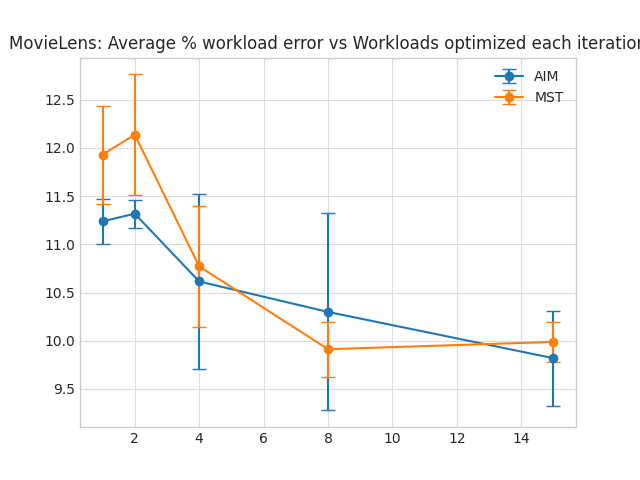}
    \includegraphics[width=0.45\linewidth]{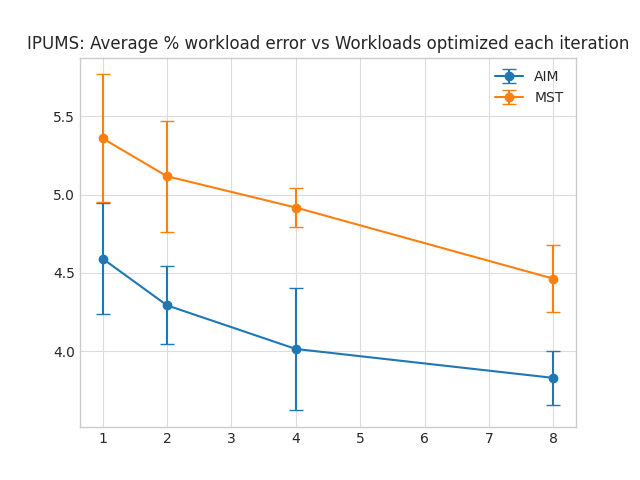}
    \caption{We demonstrate the effect of the number of workloads optimized by the relaxed optimization. As more workloads are included, the average $3$-way marginal query error reduces.}
    \label{fig:nqopt-ipums}
\end{figure}

\begin{itemize}[leftmargin=1em]
\item \textbf{Number of workloads optimized by the relaxed optimization.} 
To enhance the speed of our algorithm, we select a fixed number of workloads with the highest error each time we solve the relaxed optimization, choosing from all the workloads identified by the exponential mechanism so far. This selection strategy prioritizes computational efficiency, so we anticipate that including more workloads per step should reduce the average $3$-way marginal query error. Figure~\ref{fig:nqopt-ipums} validates this hypothesis, showing a decline in error as the number of workloads grows.
\end{itemize}

\section{Conclusion, Future Work, and Impact Statement}
\label{sec::conc}

In this paper, we investigated synthetic relational database generation with DP guarantees. We proposed an algorithm that can be combined with any preexisting single-table generation mechanisms to maintain cross-table statistical properties and referential integrity. This algorithm creates relationships between individual synthetic tables by learning a bi-adjacency matrix, which is iteratively updated through solving a quadratic program and sampling an unbiased instance to minimize approximation errors of $k$-way marginals. 
We derived rigorous utility guarantees for our algorithm and provided a PyTorch implementation suited for high-dimensional and high-volume data. Furthermore, we conducted comprehensive experiments and introduced new benchmark datasets and evaluation metrics to assess the performance and scalability of our algorithm. 
We hope our effort can inspire new research and push the frontiers of DP synthetic data towards more practical scenarios.

There are several intriguing directions that deserve further exploration. 
First, we assumed any record in the relational database must be kept private. It would be interesting to explore scenarios in which only specific tables contain personal private information. For example, in the context of education data, it may be pertinent to examine situations where only teacher and student information requires privacy-preserving, while course and department information, being often publicly available, does not. Second, our algorithm generates individual synthetic tables and uses an iterative algorithm to establish their relationship; one extension would be integrating synthetic table generation into the iterative algorithm, learning both single-table and cross-table queries simultaneously. However, achieving this may require white-box access to the generative models. Finally, while we generate synthetic relational databases with the goal of preserving low-order marginal distributions, there are other criteria worthy of exploration, such as logical consistency, temporal dynamics, and user-defined constraints. It would be valuable to understand whether these criteria can be incorporated into synthetic relational data generation while maintaining DP guarantees.

This paper focuses on privacy-preserving synthetic data generation, building upon prior efforts in DP synthetic data. We extend their application to a more practical scenario where data is stored in a relational database. This line of research may significantly impact several critical domains, such as finance, healthcare, and government, where safeguarding data privacy is paramount. The deployment of DP synthetic data can lead to more inclusive research practices, allowing organizations to share data without compromising the privacy of individuals' personal information. This, in turn, facilitates collaboration and propels advancements in research and business applications reliant on data-driven insights.

It is crucial to acknowledge potential challenges and ethical considerations associated with the deployment of synthetic data. 
Synthetic data proves to be a suitable substitute for tasks that do not necessitate absolute precision, such as data visualization, software testing, and initial model development. 
Nevertheless, there is often a trade-off between preserving the essential statistical properties of the original data and meeting privacy requirements. 
For applications with individual-level consequences, any methods or insights derived from synthetic data need to be carefully evaluated to avoid unintended consequences and biases in downstream applications. 
In summary, while our research advances the field of DP synthetic data generation, we recognize the importance of ethical considerations and the potential societal impact. By mitigating privacy concerns in data sharing, this work aspires to play a pivotal role in fostering the responsible and reliable development of advanced machine learning technologies.

\clearpage

%% file: algorithm.tex
We present our main algorithm (Algorithm~\ref{alg::adapt_alg_l2}) for generating privacy-preserving synthetic relational databases. This algorithm can be combined with any existing DP synthetic data generation mechanisms, adapting them to produce relational databases by establishing relationships among individual synthetic tables. Specifically, it learns a bi-adjacency matrix by minimizing approximation errors in terms of cross-table $k$-way marginal queries compared to the original data. Given the inherent large size of the query class, our algorithm employs an iterative approach to identify the queries with the highest approximation errors and refine the bi-adjacency matrix to reduce these errors. Notably, our algorithm is designed for efficiency and scalability, making it well-suited for high-dimensional data. It also ensures referential integrity in the generated synthetic relational data. 
We end this section by providing an overview of the rest of the paper, which elaborates on the technical details for our algorithm.

Our key observation is that cross-table marginal queries can be written as a (fractional) linear function of the bi-adjacency matrix. This formulation enables learning the bi-adjacency matrix for a synthetic relational database through an optimization problem. We present this observation formally in the following lemma.
\begin{lemma}
\label{lem::qv}
For a relational database $(\mathcal{D}_1,\mathcal{D}_2,\bB)$ and any $k$-way cross-table query $q_{(\mathcal{S}_1, \mathcal{S}_2), (\by_1, \by_2)}$, let $\ones_{(\mathcal{S}_1, \by_1)} \in \{0,1\}^{n_1}$ denote an indicator vector whose i-th element equals $1$ iff the $i$-th record in $\mathcal{D}_1$ satisfies $x_{i,j} = y_{1,j}$ for all $j\in \mathcal{S}_1$ and $\ones_{(\mathcal{S}_2, \by_2)}$ is defined in a similar manner. Let $\bq_{(\mathcal{S}_1, \mathcal{S}_2), (\by_1, \by_2)} = \operatorname{vec}(\ones_{(\mathcal{S}_1, \by_1)} \cdot \ones^T_{(\mathcal{S}_2, \by_2)})$. Then we have:
\begin{align}
    q_{(\mathcal{S}_1, \mathcal{S}_2), (\by_1, \by_2)}(\bB)
    = \frac{\bq^T_{(\mathcal{S}_1, \mathcal{S}_2), (\by_1, \by_2)} \cdot \operatorname{vec}(\bB)}{\ones^T \cdot \operatorname{vec}(\bB)}.
\end{align}
\end{lemma}
\begin{proof}
See Appendix~\ref{apx::qv}.
\end{proof}
Based on the above lemma, we write the cross-table query and its corresponding vector $\bq_{(\mathcal{S}_1, \mathcal{S}_2), (\by_1, \by_2)}$ interchangeably. We learn the bi-adjacency matrix for the synthetic database by solving the following optimization:
\begin{align}
\label{eq::l2_adj}
\min_{\substack{\bb^{\text{syn}} \in \{0,1\}^{n^{\text{syn}}_1 \cdot n^{\text{syn}}_2} \\ \bm{1}^T \bb^{\text{syn}} = m^{\text{syn}}}} \max_{\bq \in \mathcal{Q}_{\text{cross},k}} \left|\frac{1}{m^{\text{syn}}} \bq^T \bb^{\text{syn}} - a\right|,
\end{align}
where $\bb^{\text{syn}} = \operatorname{vec}(\bB^{\text{syn}})$ and $a$ is the query value computed from the original real data. 
Since the minimization over $\bb^{\text{syn}}$ in \eqref{eq::l2_adj} is a combinatorial optimization, which is inherently challenging to solve, we solve a relaxed problem by allowing $\bb^{\text{syn}} \in [0,1]^{n^{\text{syn}}_1 \cdot n^{\text{syn}}_2}$ and use an unbiased sampling algorithm to convert the obtained values back into integer values.

We present an iterative algorithm for solving the above optimization with DP guarantees. At each iteration, we apply the exponential mechanism to identify cross-table $k$-way workloads whose corresponding queries have the highest approximation errors between synthetic and real databases. Then we add isotropic Gaussian noise to their query values and update the (vectorized) bi-adjacency matrix $\bb^{\text{syn}}$ to reduce these approximation errors. Specifically, at iteration $t$, we stack all the queries reported so far into a matrix $\hat{\bQ}$ and let their noisy answers be a vector $\hat{\ba}$. To update $\bb^{\text{syn}}$, we apply projected gradient descent to solve the following quadratic program (Section~\ref{sec::proj_grad_desc}):
\begin{align}
\label{eq::relaxed_qp}
    \min_{\substack{\bb \in [0,1]^{n^{\text{syn}}_1 \cdot n^{\text{syn}}_2} \\ \bm{1}^T \bb = m^{\text{syn}}}} \left\| \frac{1}{m^{\text{syn}}} \hat{\bQ} \bb - \hat{\ba} \right\|_2^2.
\end{align}
Suppose $\bb^{\text{syn}}$ is the output from the preceding procedure, which we reshape into a matrix $\bB^{\text{syn}}$. Next, we introduce an unbiased sampling algorithm to convert each element of $\bB^{\text{syn}}$ into $\{0,1\}$ for defining the relationship between synthetic tables (Section~\ref{sec::ubs}). To ensure the scalability of our algorithm, we propose a random slicing method that allows us to generate large-scale relational databases (Section~\ref{apx::scalability}). Finally, we analyze the utility of our algorithm in preserving $k$-way marginal queries (Section~\ref{sec::utility}) and validate our algorithm through numerical experiments (Section~\ref{sec::experiments}).

We outline the above procedure in Algorithm~\ref{alg::adapt_alg_l2} and provide additional details, including our privacy budget tracking and hyper-parameters, in Section~\ref{append::main_alg}. 
Throughout the paper, we assume that the original relational database demonstrates a many-to-many relationship. In Appendix~\ref{apx::one-to-many}, we apply our algorithm to accommodate one-to-many relationships, which, as shown, are significantly easier to manage. 
We remark that iterative algorithms are a standard technique for query releasing in the DP literature \citep[see][for examples in synthetic data/graph generation]{gupta2012iterative,hardt2012simple,liu2021iterative,aydore2021differentially,mckenna2022aim} and we extend its applications to establish relationships between synthetic tables. We end this section by establishing a DP guarantee for our algorithm. 
\begin{theorem}
\label{thm::DP_guarantee}
Assume that each record in the original relational database has at most $d_{\max}$ relationships with records in the other table. We apply any DP mechanism with privacy budgets $(\epsilon_1, \delta_1)$ and $(\epsilon_2, \delta_2)$ to generate individual synthetic tables, respectively. Then, we use Algorithm
~\ref{alg::adapt_alg_l2} with privacy budget $(\epsilon_{\text{rel}}, \delta_{\text{rel}})$ to build the relationships between them. The resulting synthetic relational database satisfies $(\epsilon_1+\epsilon_2+\epsilon_{\text{rel}}, \delta_1+\delta_2+\delta_{\text{rel}})$-DP. 
\end{theorem}
\begin{proof}
See Appendix~\ref{apx::dpproof}.
\end{proof}

\begin{algorithm}[httb]
\begingroup
\small
\caption{Adapting DP mechanisms to generate relational synthetic data.}
\label{alg::adapt_alg_l2}

\begin{algorithmic}[*]

\State {\bfseries Input:} private relational database $\mathcal{B} = (\mathcal{D}_1, \mathcal{D}_2, \bB)$; privacy budget $(\epsilon_{\text{rel}}, \delta_{\text{rel}})$; individual synthetic tables $\mathcal{D}_{1}^{\text{syn}}$, $\mathcal{D}_{2}^{\text{syn}}$; workloads per iteration $K$; number of iterations $T$; number of relationship in synthetic tables $m^{\text{syn}}$

\State {\bfseries Initialize:} $\hat{\bQ} = \emptyset,\hat{\ba} = \emptyset$,  $\bb^{\text{syn}} \in [0,1]^{n_1^{\text{syn}} \cdot n_2^{\text{syn}}}$

\For{$t = 1, \cdots, T$}
\State{Select $K$ workloads via the exponential mechanism}

\State{Append all queries associated with the newly selected workloads into $\hat{\bQ}$}

\State{Add noise to the answers of the new queries and append them into $\hat{\ba}$}

\State{Solve the relaxed optimization $\tilde{\bb}^{\text{syn}} = {\arg\min}_{\substack{\bb \in [0,1]^{n_1^{\text{syn}} \cdot n_2^{\text{syn}}} \\ \bm{1}^T \bb = m^{\text{syn}}}} \lVert \frac{1}{m^{\text{syn}}}\hat{\bQ} \bb - \hat{\ba} \rVert_2^2$}\Comment{Section~\ref{sec::proj_grad_desc}}

\State {$\bB^{\text{syn}} = \operatorname{Unbiased Sampling}(\tilde{\bb}^{\text{syn}}, m^{\text{syn}})$}\Comment{Section~\ref{sec::ubs}}

\EndFor

\State {\bfseries Output:} $\mathcal{B}^{\text{syn}} = (\mathcal{D}^{\text{syn}}_1, \mathcal{D}^{\text{syn}}_2, \bB^{\text{syn}})$

\end{algorithmic}

\endgroup

\end{algorithm}

\input{sampling}

\input{slicing}

%% file: sampling.tex
\section{Solving the Relaxed Optimization}
\label{sec::proj_grad_desc}

In the previous section, we introduced our main algorithm for establishing relationships between individual synthetic tables. This algorithm updates the bi-adjacency matrix by solving a relaxed optimization and then mapping the solution back to integer values. 
In this section, we focus on solving the relaxed optimization:
\begin{equation}
\label{eq::relaxed_opt}
\begin{aligned}
    \min_{\bb \in \Omega}~f(\bb) \quad \text{where }&\ f(\bb) \defined \left\| \frac{1}{m^{\text{syn}}}\hat{\bQ} \bb - \hat{\ba} \right\|_2^2\\
    \text{and }&\ \Omega \defined \{\bb \mid 0 \leq \bb \leq 1, \ones^T \bb = m^{\text{syn}} \}\subseteq \Reals^N.
\end{aligned}
\end{equation}
Here we denote the dimension of the (vectorized) bi-adjacency matrix as $N\defined n^{\text{syn}}_1 \cdot n^{\text{syn}}_2$.

The main challenge lies in the high dimensionality of the bi-adjacency matrix (i.e., $N$ is extremely large). This makes off-the-shelf general convex quadratic program solvers either computationally expensive or unable to return solutions that satisfy the constraint set $\Omega$ in our problem. For example, running the interior point algorithm for a single iteration can take $O(N^5)$ arithmetic operations \citep{ye1989extension}, while ADMM-based solvers may fail to fully satisfy all constraints before convergence, meaning the solution might not lie within $\Omega$ \citep{schubiger2020gpu}. Ensuring all constraints are satisfied is critical since the sampling algorithm later seen in the paper requires the bi-adjacency matrix returned from the relaxed optimization meeting all constraints to guarantee that each pair of records from different synthetic tables has at most one relationship, and the total number of relationships in the synthetic database is exactly $m^{\text{syn}}$.

To address these challenges, we use the projected gradient descent algorithm to solve \eqref{eq::relaxed_opt}. This choice is motivated by our findings that both the gradient of the objective function and the projection onto the constraint set $\Omega$ can be computed in a closed form. These analytical expressions enable each iteration to run in $O(N)$ operations, significantly reducing computational costs. Moreover, the closed-form expression for the projection onto $\Omega$ ensures that the constraints are satisfied with negligible error. Finally, we establish a convergence guarantee, showing that our algorithm converges to the optimal value at a rate of $O\left(\frac{1}{T}\right)$, where $T$ is the number of iterations. This result also informs our choice of the learning rate. In the following, we provide details about our algorithm.

At iteration $t$, the projected gradient descent updates $\bb$ following the updating rule:
\begin{align}
\label{eq::pgd}
\bb_{t+1} = \Pi_\Omega \left( \bb_t - \eta \nabla f(\bb_t) \right) ,
\end{align}
where $\Pi_\Omega$ denotes the Euclidean projection onto the set $\Omega$ and $\eta$ is the step size. The gradient of the objective quadratic function has a closed-form expression:
\begin{align}
\label{eq::relaxed_grad}
    \nabla f(\bb) = \frac{2}{m^{\text{syn}}} \hat{\bQ}^T \left(\frac{1}{m^{\text{syn}}} \hat{\bQ} \bb - \hat{\ba}\right).
\end{align}
Next, we explore how to project $\bb$ (after taking the gradient descent step) onto the constraint set $\Omega$. This projection can be formulated as a quadratic program with a single sum constraint, as well as upper and lower bound constraints. Using a classical result from  \citet{pardalos1990algorithm}, we derive a closed-form expression for the projection vector in the following theorem.
\begin{theorem}
\label{thm::proj_cf}
For a given $\bb\in\Reals^N$, we define a collection of functions, indexed by $i \in [N]$, as follows:
\begin{align*}
    h_i(y) 
    \defined
    \begin{cases}
        0 \quad  &\text{if } y < -b_i,\\
        y+b_i \quad  &\text{if } -b_i \leq y \leq 1-b_i,\\
        1 \quad  &\text{if } y > 1 -b_i.
    \end{cases}
\end{align*}
Then the Euclidean projection of $\bb$ onto the set $\Omega \defined \{\bb \mid 0 \leq \bb \leq 1, \ones^T \bb = m^{\text{syn}} \}$ has a closed-form expression: 
\begin{align*}
    [\Pi_{\Omega}(\bb)]_i 
    = h_i(y^*),
\end{align*}
where $[\Pi_{\Omega}(\bb)]_i$ is the $i$-th element of $\Pi_{\Omega}(\bb)$ and $y^*$ is any real value satisfying $\sum_{i=1}^N h_i(y^*) =  m^{\text{syn}}$.
\end{theorem}
\begin{proof}
See Appendix~\ref{apx::proj_cf}.
\end{proof}
The closed-form expression for the Euclidean projection $\Pi_{\Omega}(\bb)$ relies on a \emph{single} auxiliary variable $y$. Given that $\sum_{i=1}^N h_i(y) =  m^{\text{syn}}$ is a piecewise linear, non-decreasing function, we can apply binary search over the interval $[-\max_{i\in[N]}\{b_i\} , 1-\min_{i\in[N]}\{b_i\}]$ to find a $y$ that satisfies the required equality.

Theorem~\ref{thm::proj_cf} provides several advantages for computing the projection onto $\Omega$. First, it offers an efficient algorithm with $O(N)$ runtime complexity, as it requires solving for only a single variable $y$ through the equality constraint. Moreover, the binary search approach provides full control over the solution's precision, ensuring a projection onto the constraint set with negligible error. This precision is necessary in our application since the sampling algorithm in the next section relies on the box constraint to ensure each pair of records from different synthetic tables has at most one relationship, and the sum constraint to guarantee that the total number of relationships in the synthetic database is exactly $m^{\text{syn}}$.

We can now compute the gradient of the objective function in closed form \eqref{eq::relaxed_grad} and project any point onto the constraint set, also in closed form. These two results enable an efficient implementation of using the projected gradient descent to solve the relaxed optimization problem in \eqref{eq::relaxed_opt} (see Algorithm~\ref{alg::relaxed_optimization} for a detailed implementation). Below, we establish a convergence guarantee for this algorithm and analyze its runtime complexity. The results show that our algorithm produces a weighted bi-adjacency matrix that satisfies the constraints, while achieving an objective value within $O\left(\frac{1}{T}\right)$ of the optimal solution after $T$ iterations of projected gradient descent. 
\begin{theorem}
\label{thm::convergence}
Let $\bb^{*} \in \argmin_{\bb \in \Omega} f(\bb)$ be an optimal solution of the relaxed optimization problem in~\eqref{eq::relaxed_opt}. We denote the largest singular value of $\hat{\bQ}$ by $\sigma_{\text{max}}(\hat{\bQ})$. Let $b_T$ denote the $T$-th iterate of  projected gradient descent in \eqref{eq::pgd} with a step size $\eta = \frac{1}{2} \cdot \left(\frac{m^{\text{syn}}}{\sigma_{\text{max}}(\hat{\bQ})}\right)^2$. Then we have
\begin{align*}
f(\bb_T) - f(\bb^{*}) \leq \frac{C}{T},
\end{align*}
where $C \defined 3\| \bb_0 - \bb^{*} \|_2^2/\eta + f(\bb_0) - f(\bb^{*})$ is a constant that only depends on the choice of the initialization. Additionally, the runtime complexity of the projected gradient descent as a function of $N$ is $O(N)$.
\end{theorem}
\begin{proof}
See Appendix~\ref{apx::convergence}.
\end{proof}
This convergence result suggests choosing the step size $\eta$ as $\frac{1}{2} \cdot \left(\frac{m^{\text{syn}}}{\sigma_{\text{max}}(\hat{\bQ})}\right)^2$ to achieve a convergence rate of $O\left(\frac{1}{T}\right)$. This step size depends on the largest singular value of $\hat{\bQ}$, which can be efficiently computed using the power iteration method. Since the power method only involves matrix-vector multiplications in each iteration, implementing it in PyTorch allows for GPU acceleration, significantly reducing computation time.

\begin{algorithm}[httb]
\begingroup
\small
\caption{Solving the relaxed optimization via projected gradient descent.}
\label{alg::relaxed_optimization}

\begin{algorithmic}[*]

\State {\bfseries Input:} query matrix $\hat{\bQ}$; noisy answer vector $\hat{\ba}$; number of relationships in synthetic tables $m^{\text{syn}}$; number of iterations for running projected gradient descent $T$; number of iterations for running power iteration $T_{\text{power}}$

\Function{Projection}{$\bb$, $m^{\text{syn}}$}
    
    \State Find $y^*$ by solving $\sum_{i=1}^N \max\{\min\{y^*, 1-b_i\}, -b_i \} = m^{\text{syn}} - \ones^T \bb$ \Comment{Apply binary search}

    \For{$i = 1, \cdots, N$} 

    \State $b_i^{\text{proj}} = \max\{\min\{y^*, 1-b_i\}, -b_i \} + b_i$
    
    \EndFor
    
    \State {\bfseries return} $[b_1^{\text{proj}}, \cdots, b_N^{\text{proj}}]$
\EndFunction

\State \textit{// Determine the step size:}

\State Initialize $\bv \in \mathbb{R}^N$ randomly
\For{$t = 1, \cdots, T_{\text{power}}$}
    \State $\bw = \hat{\bQ}^T \hat{\bQ}\bv$
    \State $\bv = \bw / \|\bw\|_2$
\EndFor
\State $\sigma_{\text{max}} = \|\hat{\bQ}\bv\|_2$ \Comment{Compute the largest singular value}

\State $\eta = \frac{1}{2} \left(\frac{m^{\text{syn}}}{\sigma_{\text{max}}}\right)^2$ \Comment{Compute step size according to Theorem~\ref{thm::convergence}}

\State \textit{// Run projected gradient descent:}
\State Initialize $\bb_0\in \mathbb{R}^N$ randomly
\For{$t = 1, \cdots, T$}
    \State $\nabla f(\bb_{t-1}) = \frac{2}{m^{\text{syn}}} \hat{\bQ}^T\left(\frac{1}{m^{\text{syn}}} \hat{\bQ}\bb_{t-1} - \hat{\ba}\right)$
    \State $\bb_{t} = \bb_{t-1} - \eta \nabla f(\bb_{t-1})$
    \State $\bb_{t} = \Call{Projection}{\bb_{t}, m^{\text{syn}}}$
\EndFor

\State $\tilde{\bB}^{\text{syn}} = \bb_{T}.\text{reshape}(n_1^{\text{syn}}, n_2^{\text{syn}})$

\State {\bfseries Output:} weighted bi-adjacency matrix $\tilde{\bB}^{\text{syn}}$

\end{algorithmic}
\endgroup
\end{algorithm}


\section{Unbiased Sampling}
\label{sec::ubs}

In the previous section, we applied the projected gradient descent algorithm to solve the relaxed optimization problem~\eqref{eq::relaxed_qp}. This algorithm produced a weighted bi-adjacency matrix $\tilde{\bB}^{\text{syn}} \in [0,1]^{n^{\text{syn}}_1} \times [0,1]^{n^{\text{syn}}_2}$. In this section, we convert $\tilde{\bB}^{\text{syn}}$ back into the discrete space $\{0,1\}^{n^{\text{syn}}_1} \times \{0,1\}^{n^{\text{syn}}_2}$. Our goal is to sample an unbiased\footnote{Unbiased means that $\mathbb{P}(B^{\text{syn}}_{i,j} = 1) = \tilde{B}^{\text{syn}}_{i,j}$ for all $i \in [n^{\text{syn}}_1]$ and $j \in [n^{\text{syn}}_2]$.} instance $\bB^{\text{syn}}$ from $\tilde{\bB}^{\text{syn}}$ such that the synthetic relational database has exactly $m^{\text{syn}}$ relationships (i.e., $\ones^T \operatorname{vec}(\bB^{\text{syn}}) = m^{\text{syn}}$). We will discuss why the traditional sample-and-rejection method fails to provide such an unbiased sample. Then we will introduce a novel recursive sampling algorithm, prove that it yields an unbiased sample, and analyze its runtime complexity. 
Throughout this section, we simplify our notation by denoting $\tilde{\bb} \defined \operatorname{vec}(\bB^{\text{syn}})$ and omitting the superscript $\text{syn}$ from all variables. For example, $m^{\text{syn}}$ will be denoted simply as $m$.

We formalize our objective as follows: given a vector $\tilde{\bb} \in [0,1]^{N}$ that satisfies $\bm{1}^T \tilde{\bb} = m$, we aim to sample an unbiased instance $\bb \in \{0,1\}^{N}$ such that $\mathbb{P}(b_i=1) = \tilde{b}_i$ for all $i\in [N]$ and $\bm{1}^T \bb = m$. 
If we did not have the hard constraint $\bm{1}^T \bb = m$, unbiased sampling could be achieved by independently sampling each element of $\bb$ according to the corresponding probability in $\tilde{\bb}$ (i.e., $b_i \sim \mathsf{Bernoulli}(\tilde{b}_i)$ for $i \in [N]$). However, this strategy clearly does not guarantee that $\bm{1}^T \bb = m$. This can be seen by considering a special example where $\tilde{\bb} = [\tilde{b}, \cdots, \tilde{b}]$. In this case, $\mathbb{P}(\bm{1}^T \bb = m) = \binom{N}{m}\tilde{b}^{m} (1 - \tilde{b})^{N-m} \neq 1.$

A possible solution to fix this issue is to use a sample-and-rejection method. In this approach, we repeatedly sample indices from $[N] $ according to the probability distribution $\tilde{\bb}/m$. If the current index has already been selected in a previous iteration, we reject it and continue sampling until $m$ unique indices have been chosen. These selected indices will determine the positions in $\bb$ where the value is $1$. Alas, while this approach ensures $\ones^T\bb=m$, it fails to provide an unbiased sample. We demonstrate it through the following example. 
\begin{example}
Consider $N = 3$ and $m = 2$, with $\tilde{\bb} = [\tilde{b}_1, \tilde{b}_2, \tilde{b}_3]$ such that $\tilde{b}_1 + \tilde{b}_2 + \tilde{b}_3 = 2$. There are two scenarios in which the sample-and-rejection algorithm will output the index $1$ among its two output indices: (i) the index $1$ is selected in the first iteration, or (ii) the index $2$ (or $3$) is selected first, repeatedly selected and rejected, until finally, the index $1$ is selected:
$2\underbrace{2...2}_{\text{reject}} 1$ or $3\underbrace{3...3}_{\text{reject}} 1$.
Hence, we have:
\begin{align*}
    \mathbb{P}(b_1 = 1) 
    &= \frac{\tilde{b}_1}{2} + \left(\frac{\tilde{b}_2}{2}+ \frac{\tilde{b}_2^2}{4} + \cdots \right)\frac{\tilde{b}_1}{2} + \left(\frac{\tilde{b}_3}{2}+ \frac{\tilde{b}_3^2}{4} + \cdots\right)\frac{\tilde{b}_1}{2}\\
    &= \frac{\tilde{b}_1}{2} \left(1 + \frac{\tilde{b}_2}{2-\tilde{b}_2} + \frac{\tilde{b}_3}{2-\tilde{b}_3}\right).
\end{align*}
In general, this probability is not equal to $\tilde{b}_1$, leading to a biased sample when using the sample-and-rejection algorithm.
\end{example}

\begin{figure*}[t]
\centering
\includegraphics[width=0.8\linewidth]{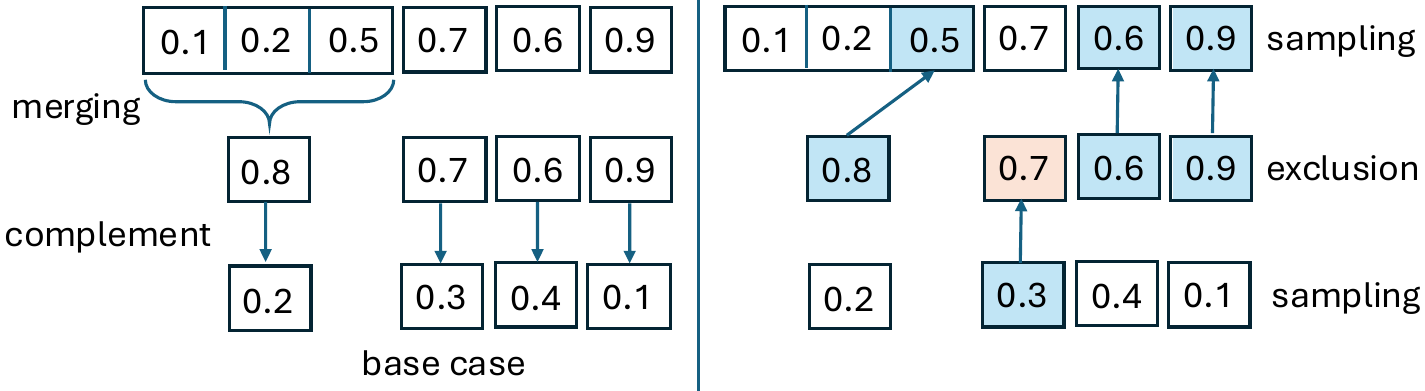}
\caption{An illustration of the function $\mathsf{UBS}$ in Algorithm~\ref{alg::rec_ubs}. Consider $\bx = [0.1, 0.2, 0.5, 0.7, 0.6, 0.9]$ and $m = 3$. In the first merging step, we partition the indices of $\bx$ into $L = 4$ groups: $\mathcal{G}[0] = \{1, 2, 3\}$, $\mathcal{G}[1] = \{4\}$, and so on. The corresponding group probabilities are $\mathsf{sum}(\bx\mid \mathcal{G}[0]) = 0.8$, $\mathsf{sum}(\bx\mid \mathcal{G}[1]) = 0.7$, etc. Next, we compute the complement of these probabilities and apply $\mathsf{UBS}([0.2, 0.3, 0.4, 0.1], 1)$ to select $L - m = 1$ group to exclude, which leads to the base case. We then sample from the probability vector $[0.2, 0.3, 0.4, 0.1]$; suppose we select group $\mathcal{G}[1]$, which is then excluded. After this, we sample exactly one index from $\mathcal{G}[0]$ according to the normalized probabilities $[0.1/0.8, 0.2/0.8, 0.5/0.8]$. Since $\mathcal{G}[2]$ and $\mathcal{G}[3]$ contain only one element each, those elements are included in the final output. The final output is the index set $\{3, 5, 6\}$.
}
\label{Fig::ubs}
\end{figure*}

We present our (recursive) unbiased sampling algorithm in Algorithm~\ref{alg::rec_ubs} and demonstrate it in Figure~\ref{Fig::ubs} with a simple example. The core of our algorithm is a function, denoted by $\mathsf{UBS}$, which takes as input a vector $\bx =[x_1, \cdots, x_N]$ where $0 \leq x_i \leq 1$ for all $i\in[N]$ and $\sum x_i = m$ for an integer $m$. It selects $m$ distinct indices from the set $[N]$ in an unbiased manner:
\begin{align*}
    \mathbb{P}(\text{index }k\text{ is selected}) = x_k \quad \forall k\in[N].
\end{align*}
This function works as follows. We sequentially partition the indices of $\bx$ into $L$ groups such that the sum of $x_i$ for each group does not exceed $1$ and cannot be increased further. Then we select $m$ groups and from each selected group, we choose exactly one index. The key idea of our algorithm is that instead of directly sampling $m$ groups, we sample $L-m$ groups according to their complement probabilities, which can be done by recursively calling the $\mathsf{UBS}$ function itself. The greedy choice of groups guarantees that the number of groups decreases exponentially in each iteration of the recursion. As a result, the algorithm has linear runtime w.r.t. $N$ (Theorem~\ref{thm::ubs}). Below, we provide more details about our algorithm (see Algorithm~\ref{alg::rec_ubs} for a pseudo-code).

\paragraph{Base cases.} We begin with three base cases. If $m = 0$, the algorithm returns an empty set. If $m = 1$, then $\bx$ already lies within a probability simplex, so we can directly sample an index according to the probability distribution $\bx$ (i.e., sample from the categorical distribution parameterized by $\bx$). If $m = N$, we know $x_i= 1$ for all $i\in [N]$, which implies that we can output the entire index set $[N]$. Otherwise, we recursively apply three steps: merge, complement, and sample.

\paragraph{Merging.} We partition the indices of $\bx$ into disjoint groups: $\mathcal{G}[0], \cdots, \mathcal{G}[L-1]$. 
These groups are formed greedily by iterating through the index set $[N]$ and sequentially adding indices to the current group. An index is added to the current group as long as the sum of $x_i$, for indices $i$ within the group, does not exceed 1. We denote by $\mathsf{sum}(\bx\mid\mathcal{G}[j])$ the sum of elements in $\bx$ whose index belongs to $\mathcal{G}[j]$:
\begin{align}
\label{eq::defn_sum_Gj}
    \mathsf{sum}(\bx\mid \mathcal{G}[j]) \defined \sum_{i \in \mathcal{G}[j]} x_i.
\end{align}
Based on our construction, we have $\mathcal{G}[j] = \{z_j, z_j + 1, \cdots, z_{j+1}-1\} \subseteq [N]$, $1 = z_0 < z_1 < \cdots < z_{L} = N+1$, and $0\leq \groupsum{j} \leq 1$.

\paragraph{Complement.}
We aim to select $m$ groups from $\mathcal{G}[0], \cdots, \mathcal{G}[L-1]$ such that the sampling process is unbiased:
\begin{align*}
    \mathbb{P}(\text{group } \mathcal{G}[j] \text{ is selected}) = \mathsf{sum}(\bx\mid \mathcal{G}[j]).
\end{align*}
At first glance, one might attempt to solve this problem by applying $\mathsf{UBS}(\bp, m)$, where $\bp = [p_0, \dots, p_{L-1}]$ with $p_j = \mathsf{sum}(\bx \mid \mathcal{G}[j])$. However, this approach fails because $\bp$ cannot be further merged, causing the recursive algorithm to enter an infinite loop. Our key idea is to apply $\mathsf{UBS}( \ones - \bp, L - m)$ to select $L - m$ groups to exclude. In other words, instead of directly selecting $m$ groups from $\mathcal{G}[0], \dots, \mathcal{G}[L-1]$, we reframe the problem: how can we select $L - m$ groups based on the complement probabilities? These groups will be ruled out, and the remaining $m$ groups will be selected.

\paragraph{Sampling.} Suppose $\mathcal{G}[j]$ is one of the $m$ remaining non-excluded groups. From $\mathcal{G}[j]$, we sample exactly one index $i$ with probability proportional to $x_i$:
\begin{align*}
    \mathbb{P}(\text{$i$ is sampled} \mid \text{group $\mathcal{G}[j]$ is selected}) = \frac{x_i}{\groupsum{j}}\quad \text{ for } i \in \mathcal{G}[j].
\end{align*}
We return the set of indices selected from all $m$ non-excluded groups.

\begin{algorithm}[httb]
\begingroup
\small
\caption{A recursive algorithm for unbiased sampling.}
\label{alg::rec_ubs}

\begin{algorithmic}[*]

\State {\bfseries Input:} weighted bi-adjacency matrix $\tilde{\bB}^{\text{syn}} \in [0,1]^{n^{\text{syn}}_1} \times [0,1]^{n^{\text{syn}}_2}$, number of relationship in synthetic tables $m^{\text{syn}}$

\Function{UBS}{$\bx$, $m$}
    \State Assert $\ones^T \bx == m$
    \State $N = \text{length}(\bx)$
    \If{$m == 0$}
        \State {\bfseries return} $\{\}$
    \ElsIf{$m == 1$}
        \State Draw $k$ from $\{1, \cdots, N\}$ according to the probability distribution $\bx$
        \State {\bfseries return} $\{k\}$
    \ElsIf{$m == N$}
        \State {\bfseries return} $\{1, \cdots, N\}$
    \EndIf
    \State \textit{// Merging procedure:}
    \State $\mathcal{G} = [\{\}]$ \Comment{List of groups, where groups are sets of indices}
    \State $i = 0$
    \While{$i < N$}
        \If{$\mathsf{sum}(\bx\mid\mathcal{G}[-1]) + x_i > 1$}
            \State Append an empty set to $\mathcal{G}$ \qquad \Comment{Finish the current group}
        \Else
            \State Add $i$ to the set $\mathcal{G}[-1]$
        \EndIf
        \State $i = i + 1$
    \EndWhile
    \State \textit{// Complement procedure and recursive step:}
    \State $L = \text{length}(\mathcal{G})$
    \State $\bp' = [1 - \mathsf{sum}(\bx\mid\mathcal{G}[i]) \ \text{ for }\  i = 0,\cdots, L-1]$
    \State $\mathcal{J} = \Call{UBS}{\bp', L - m}$
    \State \textit{// Sampling procedure:}
    \State  $\mathcal{O} = \{\}$
    \For{$i = 1,\cdots, L$}
        \If{$i \notin \mathcal{J}$}
            
            \State $\bq = [x[j] \ \text{ for }\  j \in \mathcal{G}[i]]$ and $\bq = \bq/(\ones^T \bq)$
            \State Draw $s$ from $\mathcal{G}[i]$ according to the probability distribution $\bq$
            \State Add $s$ to $\mathcal{O}$
        \EndIf
    \EndFor
    
    \State {\bfseries return} $\mathcal{O}$
\EndFunction

\State $\tilde{\bb}^{\text{syn}} = \tilde{\bB}^{\text{syn}}.\text{flatten}()$

\State $\mathcal{O} = \Call{UBS}{\tilde{\bb}^{\text{syn}},m^{\text{syn}}}$

\State $\bb^{\text{syn}} = [1 \ \text{ if }\ i \in \mathcal{O} \ \text{ else } 0 \ \text{ for }\ i = 1,\cdots, n_1^{\text{syn}}\cdot n_2^{\text{syn}}]$

\State $\bB^{\text{syn}} = \bb^{\text{syn}}.\text{reshape}(n_1^{\text{syn}}, n_2^{\text{syn}})$

\State {\bfseries Output:} unweighted bi-adjacency matrix $\bB^{\text{syn}}$

\end{algorithmic}
\endgroup
\end{algorithm}

\begin{theorem}
\label{thm::ubs}
The function $\mathsf{UBS}$ in Algorithm~\ref{alg::rec_ubs} outputs an unbiased sample. Additionally, the runtime of Algorithm~\ref{alg::rec_ubs} is $O(N)$ where $N = n^{\text{syn}}_1 \cdot n^{\text{syn}}_2$.
\end{theorem}
\begin{proof}
See Appendix~\ref{append::samplingproof}.
\end{proof}

An unbiased sampling approach offers significant benefits over alternative methods. Suppose Algorithm~\ref{alg::rec_ubs} takes a weighted bi-adjacency matrix $\tilde{\bB}^{\text{syn}}$ as input and produces $\bB^{\text{syn}}$ as output. The unbiased property ensures that $\EE{\bB^{\text{syn}}} = \tilde{\bB}^{\text{syn}}$. Given that $k$-way cross-table queries are linear functions of the bi-adjacency matrix, computing these queries from $\bB^{\text{syn}}$ provides an unbiased estimate of the corresponding queries computed from $\tilde{\bB}^{\text{syn}}$. Specifically, Lemma~\ref{lem::qv} implies that, for any query vector $\bq$, its value computed from a relational database with bi-adjacency matrix $\bB$ is $\frac{1}{m^{\text{syn}}} \bq^T \operatorname{vec}(\bB)$. Thus, we have 
\begin{align*}
\EE{\frac{1}{m^{\text{syn}}} \bq^T \operatorname{vec}(\bB^{\text{syn}})} = \frac{1}{m^{\text{syn}}}\bq^T \operatorname{vec}(\tilde{\bB}^{\text{syn}}).
\end{align*}
In other words, as long as the relaxed optimization in Section~\ref{sec::proj_grad_desc} yields an accurate solution, the unbiased sample can maintain accuracy (on average) in estimating marginal queries.

Next, we provide a high-probability bound to control the deviation of $\frac{1}{m^{\text{syn}}}\bq^T \operatorname{vec}(\bB^{\text{syn}})$ from its expected value for specific instances generated by Algorithm~\ref{alg::rec_ubs}. Typically, this can be achieved using Hoeffding's inequality \citep{hoeffding63ineqbrv} or related concentration bounds. However, standard Hoeffding bounds assume independent random variables—a condition not met in our sampling problem due to the sum constraint $\ones^T \operatorname{vec}(\bB^{\text{syn}}) = m^{\text{syn}}$. To resolve this issue, we prove that the elements of $\bB^{\text{syn}}$ are negatively associated random variables \citep{Joagdev1983NegativeAO,dubhashi1998NA}, enabling us to apply Hoeffding's bound effectively. A key step of our proof is to show that the merging and complement steps in our algorithm preserve the negative association property. We formally state this result in the following theorem.
\begin{theorem}
\label{thm::concentration}
Suppose Algorithm~\ref{alg::rec_ubs} takes a weighted bi-adjacency matrix $\tilde{\bB}^{\text{syn}}$ as input and outputs $\bB^{\text{syn}}$. Let $\mathcal{Q}$ represent a set of $k$-way cross-table queries of interest. 
Then, with probability at least $1-\beta$, any query vector $\bq \in \mathcal{Q}$ satisfies
\begin{align*}
    \left| \frac{1}{m^{\text{syn}}} \bq^T \operatorname{vec}(\bB^{\text{syn}}) -  \frac{1}{m^{\text{syn}}}\bq^T \operatorname{vec}(\tilde{\bB}^{\text{syn}})\right| \leq \frac{1}{m^{\text{syn}}} \cdot \sqrt{\frac{N}{2} \cdot \log \frac{2 | \mathcal{Q}|}{\beta}},
\end{align*}
where $N = n^{\text{syn}}_1 \cdot n^{\text{syn}}_2$.
\end{theorem}
\begin{proof}
    See Appendix~\ref{apx::NA}.
\end{proof}
The results in the above theorem apply not only to $k$-way marginal queries but also to any statistical queries that can be expressed as a linear function of the bi-adjacency matrix.
Additionally, we remark that \citet{srinivasan2001Sampling} also studied the problem of sampling $N$ indicator random variables with a sum constrained to $m^{\text{syn}}$ and proposed an unbiased sampling method. Although their algorithm and ours both have a worst-case time complexity of $O(N)$, empirical comparisons show that our algorithm runs significantly faster when the ratio $m^\text{syn}/N$ is small. This improvement arises because, in our Algorithm~\ref{alg::rec_ubs}, only the initial merging step requires $O(N)$ operations; thereafter, the probability vector’s dimensionality reduces to $\leq 2m^{\text{syn}}+1$, yielding $O(m^{\text{syn}})$ operations in subsequent steps (see Appendix~\ref{append::samplingproof} for details). In contrast, the runtime of the algorithm in \citet{srinivasan2001Sampling} does not decrease with $m^{\text{syn}}$. In practice, $m^\text{syn}/N$ is often quite small (e.g., approximately $4.2 \times 10^{-4}$ in the \texttt{MovieLens} dataset), making our algorithm significantly faster than that of \citet{srinivasan2001Sampling}.



%% file: slicing.tex
\section{Extension for Improving Scalability}
\label{apx::scalability}

\begin{figure*}[t]
\centering
\includegraphics[width=\textwidth]{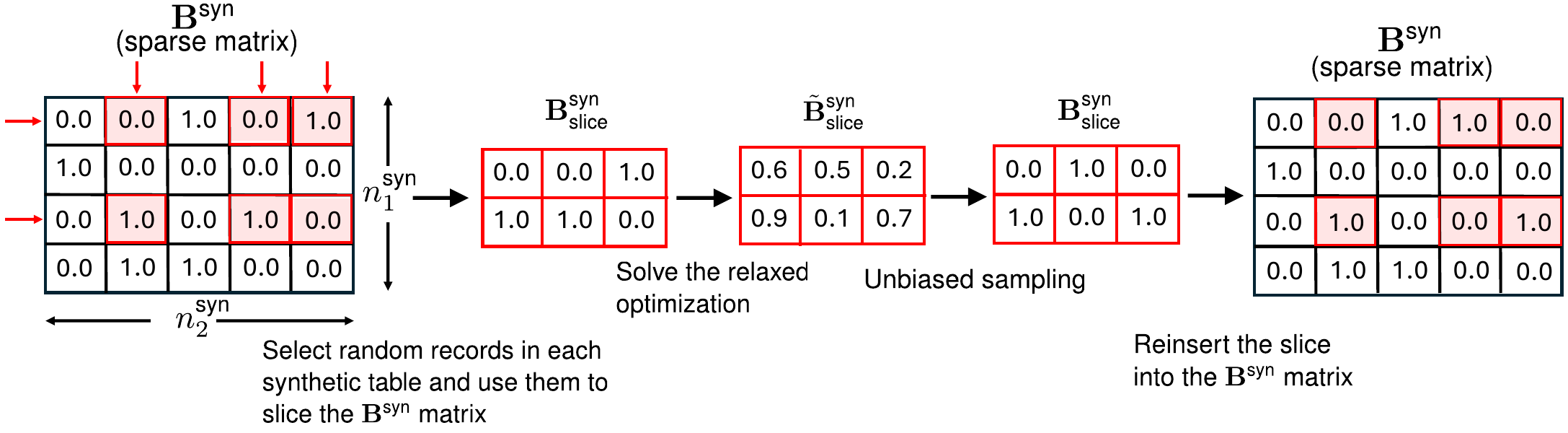}
\vspace{-20pt}
\caption{Illustration of how the random slicing heuristic enhances the scalability of the main algorithm. A full version of the main algorithm, incorporating the random slicing heuristic, is provided in Appendix~\ref{append::main_alg}.
} 
\label{fig:diagram_algorithm}
\end{figure*}

In the previous sections, we presented a projected gradient descent algorithm for learning a relaxed bi-adjacency matrix, along with an unbiased sampling algorithm to convert this matrix back to discrete space. However, the runtime and memory requirements of both algorithms inherently depend on $n_1^{\text{syn}} \times n_2^{\text{syn}}$, which is the dimension of the bi-adjacency matrix $\bB^{\text{syn}}$ for the synthetic database. In practice, this dimension can be extremely large. For example, in our experiments with the \texttt{MovieLens} dataset \citep{harper2015movielens}, the user and movie tables contain $6,040$ and $3,900$ records, respectively. To generate a synthetic relational database with the same number of entries, $n_1^{\text{syn}} \times n_2^{\text{syn}}$ would be nearly \emph{$24$ million}. This high dimensionality makes it impractical to learn a bi-adjacency matrix that accurately captures the noisy observations derived from real data. To address this challenge, in this section, we introduce a heuristic---random slicing---that can facilitate the synthesis of large-scale relational databases (see Figure~\ref{fig:diagram_algorithm} for an illustration and Appendix~\ref{append::main_alg} for its integration into our main algorithm).

The key idea of our heuristic is to explore the sparsity of the bi-adjacency matrix $\bB^{\text{syn}}$. Specifically, we observe that the non-zero entries of $\bB^{\text{syn}}$ typically scale as $O(n_1^{\text{syn}} + n_2^{\text{syn}})$. For example, in the \texttt{MovieLens} dataset, users tend to rate only a small subset of movies, and in the student-teacher scenario, each student generally enrolls in at most $10$ courses, each taught by a different teacher. In such cases, the number of non-zero entries in $\bB^{\text{syn}}$ (denoted $m^{\text{syn}}$) increases linearly with the size of the user (or student) synthetic table. Based on this, we store $\bB^{\text{syn}}$ as a sparse matrix and update a subset of its elements at each iteration of Algorithm~\ref{alg::adapt_alg_l2}.

Our heuristic works as follows. At each iteration of Algorithm~\ref{alg::adapt_alg_l2}, we randomly select a fraction $\tau \in (0,1)$ of records from each synthetic table and slice $\bB^{\text{syn}}$ to include only the relationships among the selected records in each table. We denote the sliced bi-adjacency matrix as $\bB^{\text{syn}}_{\text{slice}}$. We then solve a similar relaxed optimization using the projected gradient descent algorithm (Section~\ref{sec::proj_grad_desc}) to optimize the relationships among these selected records, followed by applying the unbiased sampling algorithm (Section~\ref{sec::ubs}) to discretize the sliced bi-adjacency matrix. Finally, we re-insert these newly learned relationships into $\bB^{\text{syn}}$. This approach reduces the matrix size for the projected gradient descent and the unbiased sampling algorithms from $n_1^{\text{syn}} n_2^{\text{syn}}$ to $\tau^2 n_1^{\text{syn}} n_2^{\text{syn}}$. 
Since these algorithms have linear time and space complexity w.r.t. the matrix size, their complexity is reduced to $O(\tau^2 n_1^{\text{syn}} n_2^{\text{syn}})$, leading to runtime and memory savings. 
%

\section{Utility Guarantees}
\label{sec::utility}

We present a utility theorem that provides an upper bound on the approximation error of cross-table queries using synthetic relational data generated by Algorithm~\ref{alg::adapt_alg_l2}. 
We begin by defining how to quantify approximation errors for a synthetic relational database $(\mathcal{D}_1^{\text{syn}},\mathcal{D}_2^{\text{syn}},\bB^{\text{syn}})$. Suppose this synthetic database has $m^\text{syn}$ relationships. We stack the query vectors of interest as a matrix $\hat{\bQ}$. Lemma~\ref{lem::qv} shows that the query answers derived from this synthetic database can be expressed as $\frac{1}{m^\text{syn}} \hat{\bQ} \operatorname{vec}(\bB^{\text{syn}})$. We now introduce the following definition to measure how accurately $k$-way cross-table queries computed from the synthetic database approximate those obtained from the original data. In this definition, we allow $\bB^{\text{syn}}$ to be a weighted bi-adjacency matrix.
\begin{definition} 
\label{defn::mse}
Let $(\mathcal{D}_1^{\text{syn}},\mathcal{D}_2^{\text{syn}},\bB^{\text{syn}})$ be a synthetic relational database such that $m^{\text{syn}} = \bm{1}^T \operatorname{vec}(\bB^{\text{syn}})$. We stack all the $k$-way cross-table query vectors as $\hat{\bQ}$. Denote by $\ba$ the true answers to these queries computed from the original database. We define the (per-workload) mean squared error as
\begin{align}
    \mse(\mathcal{D}_1^{\text{syn}}, \mathcal{D}_2^{\text{syn}}, \bB^{\text{syn}}) 
    \defined \frac{1}{|\mathcal{W}_{\text{cross}, k}|} \cdot \left\| \frac{1}{m^\text{syn}} 
    \hat{\bQ} \operatorname{vec}(\bB^{\text{syn}}) - \ba \right\|_2^2,
\end{align}
where $|\mathcal{W}_{\text{cross}, k}|$ denotes the total number of $k$-way cross-table marginal workloads.
\end{definition}

We now present our utility theorem. It provides an upper bound for the mean squared error in approximating cross-table queries based on outputs from our algorithm. This bound depends on the error associated with the optimal bi-adjacency matrix for the given synthetic tables (i.e., the best achievable error without privacy constraints) and an additional term that approaches zero as the size of the original database increases. The proof techniques are based on classical tools from convex geometry and concentration inequalities.

\begin{theorem}
\label{thm::Projection_utility}
Consider running a single iteration of Algorithm~\ref{alg::adapt_alg_l2} (without unbiased sampling) to establish relationships between any individual synthetic tables $\mathcal{D}_1^{\text{syn}}$ and $\mathcal{D}_2^{\text{syn}}$. The output is a bi-adjacency matrix $\bB^{\text{syn}}$. Let $\bB^{*}$ denote the optimal weighted bi-adjacency matrix in minimizing the mean squared error (Definition~\ref{defn::mse}) when no privacy constraints are enforced. 
With high probability, we have
\begin{align*}
    \mse(\mathcal{D}_1^{\text{syn}}, \mathcal{D}_2^{\text{syn}}, \bB^{\text{syn}}) 
    \leq 8 \cdot \mse(\mathcal{D}_1^{\text{syn}}, \mathcal{D}_2^{\text{syn}}, \bB^*) + \tilde{O}\left(\frac{d_{\text{max}}}{m\sqrt{\rho_{\text{rel}}}}\right),
\end{align*}
where $\rho_{\text{rel}} = \left(\sqrt{\epsilon_{\text{rel}} + \operatorname{log} \frac{1}{\delta_{\text{rel}}}} -  \sqrt{\operatorname{log} \frac{1}{\delta_{\text{rel}}}}\right)^2$, $m$ is the number of relationships in the original database, and $\tilde{O}(\cdot)$ ignores logarithmic factors.
\end{theorem}
\begin{proof}
See Appendix~\ref{apx::utilityproof}.
\end{proof}
The theorem above establishes an upper bound on the (per-workload) squared error when estimating cross-table queries using synthetic data generated from Algorithm~\ref{alg::adapt_alg_l2}. This upper bound consists of the optimal error without privacy constraints plus an additional error term. The optimal error only hinges on the quality of the individual synthetic tables, determined by the privacy mechanism used for generating these tables. In a hypothetical scenario where the synthetic tables are identical to the original tables, this term would be reduced to zero. 
On the other hand, the additional error term depends on three factors: $m$ (the number of relationships in the original database), $d_{\max}$ (the maximum number of relationships a record in the original database can have), and $\rho_{\text{rel}}$ (privacy parameter). 
With fixed $d_{\max}$ and $\rho_{\text{rel}}$, as the number of relationships increases, the additional error term can be reduced to zero. 
Finally, we note that existing utility theorems \citep[e.g.,][]{liu2023generating,aydore2021differentially} often rely on the size of data domain $|\mathcal{X}_i|$---that is, the number of distinct values for each categorical feature. This quantity could be large, particularly if data are discretized into numerous bins. 
We eliminate this dependence in our utility theorem by leveraging the ``marginal trick'' \citep[see Appendix~D.4 in][]{liu2021iterative} in our algorithm.

%% file: appendix.tex
\section{Background on Differential Privacy (DP)}
\label{apx::dp}

We recall the concept of (zero) concentrated differential privacy \citep{bun2016concentrated,steinke2022composition}. It will be used in the proof of Theorem~\ref{thm::DP_guarantee} for establishing DP guarantees for our algorithm.

\begin{definition}
(Zero concentrated differential privacy (zCDP)). A randomized mechanism $\mathcal{M}$ satisfies $\rho$-zero Concentrated Differential Privacy (zCDP) if for all pairs of neighboring relational databases $\mathcal{B}, \mathcal{B}'$, and for all $\alpha \in(1, \infty)$ :
$$
\mathbb{D}_\alpha\left(\mathcal{M}(\mathcal{B}), \mathcal{M}\left(\mathcal{B}\right)\right) \leq \rho \alpha,
$$
where $\mathbb{D}_\alpha\left(\mathcal{M}(\mathcal{B}), \mathcal{M}\left(\mathcal{B}'\right)\right)$ denotes the $\alpha$-Renyi divergence between the distributions of $\mathcal{M}(\mathcal{B})$ and $\mathcal{M}\left(\mathcal{B}'\right)$.
\end{definition}
zCDP has nice composition and post-processing properties. 
\begin{lemma}
\label{lem::comp-zcdp} (Adaptive composition). 
Let $\mathcal{M}_1, \cdots,\mathcal{M}_T$ be $\rho_1, \cdots, \rho_T$-zCDP, respectively. Suppose $\mathcal{M}$ is an adaptive composition of $\mathcal{M}_1,\cdots,\mathcal{M}_T$. Then $\mathcal{M}$ satisfies $(\sum \rho_i)$-zCDP.
\end{lemma}

\begin{lemma}
\label{lem::pp-zcdp}
(Post-processing). 
Let $\mathcal{M}$ be a $\rho$-zCDP mechanism. Suppose $f$ is any (possibly randomized) function. Then $\mathcal{M}^{\prime}(\mathcal{B})=f(\mathcal{M}(\mathcal{B}))$ satisfies $\rho$-zCDP.
\end{lemma}

\begin{lemma}
\label{lem::approx-zcdp}
(Relation between zCDP and DP).
If $\mathcal{M}$ satisfies $\rho$-zCDP, then it also satisfies $(\rho+2 \sqrt{\rho \log (1 / \delta)}, \delta)$-DP for any $\delta>0$.
\end{lemma}

Next, we recall two basic privacy mechanisms: Gaussian and exponential mechanisms. 
To start with, we define the $L_p$ sensitivity of a function $f$ as  
\begin{align}
    \Delta_p(f) \defined \sup_{\mathcal{B} \sim \mathcal{B}'} \|f(\mathcal{B}) - f(\mathcal{B}') \|_p, 
\end{align}
where $\mathcal{B} \sim \mathcal{B}'$ are two neighboring databases. 
\begin{lemma}
\label{lem::Gaussian}
(Gaussian Mechanism). Given a function $f$, the Gaussian mechanism is defined as $\mathcal{M}(\mathcal{B}) = f(\mathcal{B}) + \sigma \Delta_2(f) N$ where $N$ is a random vector drawn from $\mathcal{N}(\bm{0}, \bI)$. Then $\mathcal{M}$ satisfies $\frac{1}{2\sigma^2}$-zCDP.
\end{lemma}
Finally, we recall the exponential mechanism. For a $\mathsf{score}$ function and a set $\mathcal{O}$, we define its sensitivity by
\begin{align}
    \Delta(\mathsf{score})
    \defined \sup_{\mathcal{B}\sim \mathcal{B}'} \max_{o\in \mathcal{O}} |\mathsf{score}(\mathcal{B}, o) -  \mathsf{score}(\mathcal{B}', o)|.
\end{align}

\begin{lemma}
\label{lem::exponential}
(Exponential Mechanism). Given a $\mathsf{score}$ function and the set $\mathcal{O}$ of all possible outputs, the exponential mechanism $\mathcal{M}$ is a randomized mechanism defined by
\begin{align}
    \mathbb{P}(\mathcal{M}(\mathcal{B}) = o) 
    \propto \exp\left( \frac{\epsilon}{2\Delta(\mathsf{score})} \mathsf{score}(\mathcal{B}, o)\right),\quad \forall o \in \mathcal{O}.
\end{align}
Then $\mathcal{M}$ satisfies $\frac{\epsilon^2}{8}$-zCDP.
\end{lemma}

\section{More Details about Algorithm~\ref{alg::adapt_alg_l2}}
\label{append::main_alg}

In this section, we outline the implementation details of our main algorithm (Algorithm~\ref{alg::adapt_alg_l2}). The full version is presented in Algorithm~\ref{alg::alg_sparse_large} and illustrated in Figure~\ref{fig:diagram_algorithm}. Recall that our algorithm selects worst-case workloads using the exponential mechanism and computes noisy observations of these workloads from real data using the Gaussian mechanism. Details on these mechanisms can be found in Algorithms~\ref{alg::exponential} and \ref{alg::Gaussian}, along with their respective privacy budget allocations. Additionally, we incorporate the random slicing mechanism (Section~\ref{apx::scalability}) to enhance scalability for generating large-scale synthetic relational databases.

To implement the random slicing mechanism, we begin by storing the bi-adjacency matrix as a sparse matrix $\bB^{\text{syn}} \in \{0, 1\}^{n_1^{\text{syn}}} \times \{0, 1\}^{n_2^{\text{syn}}}$. We introduce a parameter $\tau < 1$, which specifies the fraction of relations to be learned in each iteration. Next, we define the slicing operation, which will be used to extract elements from $\bB^{\text{syn}}$ before running the projected gradient descent and unbiased sampling algorithms. 
\begin{definition}
\label{defn::slice}
Given a matrix $\bM \in \Reals^{n_1\times n_2}$ and vectors $\bv_1 \in \{0, 1\}^{n_1}$, $\bv_2 \in \{0, 1\}^{n_2}$, we define the slicing operation as $\left[\bv^1\right]\bM\left[\bv^2\right]$. This operation yields a matrix with $\ones^T {\bv^{\text{slice}}_1}$ rows and $\ones^T{\bv^{\text{slice}}_2}$ columns. The $(i, j)$ entry of this matrix corresponds to the $(v^1_i, v^2_j)$ entry of $\bM$, where $v^1_i$ is defined as the $i$-th nonzero entry of $\bv^1$ (similarly for $v^2_j$). If only the columns are sliced, we write $\bM \left[\bv^2\right]$.
\end{definition}

\begin{algorithm}[h]
\begingroup
\small
\caption{Full Algorithm \ref{alg::adapt_alg_l2} (detailed version with scalability enhancements).}
\label{alg::alg_sparse_large}

\begin{algorithmic}[*]

\State {\bfseries Input:} private relational database $\mathcal{B} = (\mathcal{D}_1, \mathcal{D}_2, \bB)$; privacy budget $(\epsilon_{\text{rel}}, \delta_{\text{rel}})$; individual synthetic tables $\mathcal{D}_{1}^{\text{syn}}$, $\mathcal{D}_{2}^{\text{syn}}$; workloads per iteration $K$; number of iterations $T$; number of relationship in synthetic tables $m^{\text{syn}}$; slicing fraction $\tau$; hyper-parameter balancing Gaussian and exponential mechanisms $\alpha$

\State {\bfseries Initialize:} $\hat{\bQ} = \emptyset,\hat{\ba} = \emptyset$, $n_1^{\text{syn}}$ by $n_2^{\text{syn}}$ sparse matrix $\bB^{\text{syn}}$ with entries $\in \{0,1\}$

\State Convert $(\epsilon_{\text{rel}}, \delta_{\text{rel}})$-DP into $\rho_{\text{rel}}$-zCDP and compute a privacy parameter $\epsilon_0 = \sqrt{\frac{2\rho_{\text{rel}}}{KT}}$

\For{$t = 1, \cdots, T$}
\State{Select $K$ workloads via the exponential mechanism}\Comment{Algorithm~\ref{alg::exponential}}

\State{Append all queries associated with the newly selected workloads into $\hat{\bQ}$}

\State{Add noise to the answers of the new queries and append them into $\hat{\ba}$}\Comment{Algorithm~\ref{alg::Gaussian}}

\State{Randomly choose $\bv^{\text{slice}}_1 \in \{0, 1\}^{n_1^{\text{syn}}}$, $\bv^{\text{slice}}_2 \in \{0, 1\}^{n_2^{\text{syn}}}$ s.t. $\mathsf{sum}({\bv^{\text{slice}}_i}) = \floor{\tau n_i^{\text{syn}}}$ for $i \in \{1, 2\}$}

\State{Set $\bB^{\text{syn}}_{\text{slice}} = [\bv^{\text{slice}}_1]\bB^{\text{syn}}[\bv^{\text{slice}}_2]$ and $\bQ_{\text{slice}} = \bQ[\operatorname{vec}(\bv^{\text{slice}}_1\cdot (\bv^{\text{slice}}_2.\text{transpose}))]$}\Comment{Definition~\ref{defn::slice}}

\State{Compute $m_{\text{slice}} = \mathsf{sum}(\bB^{\text{syn}}_{\text{slice}})$}

\State{Solve the relaxed optimization:}\Comment{Algorithm~\ref{alg::relaxed_optimization}}
\begin{align*}
    \tilde{\bB}^{\text{syn}}_{\text{slice}} = \argmin_{\substack{B_{i,j} \in [0,1] \\ \mathsf{sum}(\bB_{\text{slice}}^{\text{syn}}) = m_{\text{slice}}}} \lVert \frac{1}{m_{\text{slice}}}\bQ_{\text{slice}} \mathsf{flatten}(\bB_{\text{slice}}^{\text{syn}}) - \hat{\ba} \rVert_2^2
\end{align*}

\State {$\bB^{\text{syn}}_{\text{slice}} = \operatorname{UnbiasedSampling}(\mathsf{flatten}(\tilde{\bB}^{\text{syn}}_{\text{slice}}), m_{\text{slice}})$}\Comment{Algorithm~\ref{alg::rec_ubs}}

\State{$[\bv^{\text{slice}}_1]\bB^{\text{syn}}[\bv^{\text{slice}}_2] = \bB^{\text{syn}}_{\text{slice}}$}\Comment{Re-insert $\bB^{\text{syn}}_{\text{slice}}$ into $\bB^{\text{syn}}$}

\EndFor

\State {\bfseries Output:} $\mathcal{B}^{\text{syn}} = (\mathcal{D}^{\text{syn}}_1, \mathcal{D}^{\text{syn}}_2, \bB^{\text{syn}})$

\end{algorithmic}

\endgroup

\end{algorithm}

\begin{algorithm}[httb]
\begingroup
\small
\caption{Exponential mechanism for workload selection.}
\label{alg::exponential}

\begin{algorithmic}[*]

\State {\bfseries Input:} original private relational database $\mathcal{B} = (\mathcal{D}_1, \mathcal{D}_2, \bB)$; synthetic relational database $\mathcal{B}^{\text{syn}} = (\mathcal{D}_{1}^{\text{syn}}, \mathcal{D}_{2}^{\text{syn}}, \bB^{\text{syn}})$; workloads per iteration $K$; all $k$-way cross-table workloads $\{(\mathcal{S}_1^{1}, \mathcal{S}_2^{1}), \cdots, (\mathcal{S}_1^{W}, \mathcal{S}_2^{W})\}$; workloads that have already been selected $\mathcal{T}_{\text{selected}}$; hyper-parameter balancing Gaussian and exponential mechanisms $\alpha$; privacy parameter $\epsilon_0$

\State {\bfseries Initialize:} $\mathcal{O} = \emptyset$ and $\mathcal{W} = \mathsf{Subsample}\left(\{(\mathcal{S}_1^{1}, \mathcal{S}_2^{1}), \cdots, (\mathcal{S}_1^{W}, \mathcal{S}_2^{W})\} \backslash \mathcal{T}_{\text{selected}} \right)$\Comment{Save runtime}

\For{$w \in \mathcal{W}$}

\State $\mathsf{score}(w) = \frac{1}{2} \|P_{w}(\mathcal{B}) - P_{w}(\mathcal{B}^{\text{syn}})\|_1$\Comment{\parbox[t]{.5\linewidth}{total variance distance between marginal distributions of $\mathcal{B}$ and $\mathcal{B}^{\text{syn}}$ on workload $w$}}

\EndFor

\For{$t = 1, \cdots, K$}

\State $\bp$ = [$\mathsf{score}(w)$ for $w\in\mathcal{W}$]\Comment{Scores of workloads that have not been selected}

\State $\bp = \mathsf{softmax}\left(\sqrt{ \alpha}\epsilon_0 \frac{m}{d_{\max}} \bp\right)$\Comment{$m = \ones^T \cdot \operatorname{vec}(\bB)$ and $d_{\max}$ is defined in Section~\ref{subsec::bipar_graph}}

\State Sample a workload $w$ according to the probability vector $\bp$

\State $\mathcal{O} = \mathcal{O} \cup {w}$ and $\mathcal{W} = \mathcal{W} \backslash \{w\}$

\EndFor

\State $\mathcal{T}_{\text{selected}} = \mathcal{T}_{\text{selected}} \cup \mathcal{O}$

\State {\bfseries Output:} $\mathcal{O}$

\end{algorithmic}

\endgroup

\end{algorithm}

\begin{algorithm}[httb]
\begingroup
\small
\caption{Gaussian mechanism for noisy answer computation.}
\label{alg::Gaussian}

\begin{algorithmic}[*]

\State {\bfseries Input:} (empirical) marginal distributions of $\mathcal{B}$ on $K$ newly selected workloads $\{\bp_1, \cdots, \bp_K \}$; hyper-parameter balancing Gaussian and exponential mechanisms $\alpha$; privacy parameter $\epsilon_0$

\For{$t = 1, \cdots, K$}

\State{Assert $\operatorname{sum}(\bp_t) == 1$ and $\bp_t \geq 0$ }

\State $\hat{\ba}_t = \bp_t + \left( \frac{\sqrt{2} d_{\max}}{m \sqrt{(1-\alpha)} \epsilon_0} \right) \mathcal{N}(\bm{0},  \bI)$\Comment{Add Gaussian noise to each probability vector}

\EndFor

\State {\bfseries Output:} $\{\hat{\ba}_1, \cdots, \hat{\ba}_K\}$

\end{algorithmic}

\endgroup

\end{algorithm}

\section{One-to-Many Relationship}
\label{apx::one-to-many}

In the main body of the paper, we focused on many-to-many relationship. In some cases, the original database may exhibit a one-to-many relationship. For instance, in the context of education data (e.g., student and school data), each school can have multiple students, but each student belongs to only one school. When generating a synthetic database, it is crucial to preserve this relationship. In this section, we apply our algorithm to this scenario. As shown, by incorporating this additional assumption---that records in one table have exactly one relationship with the other table---our algorithm becomes significantly simpler and operates much more efficiently.

Consider that each record in the first table of the original database has only one corresponding relationship (e.g., the student table from the example above). We denote the rows of the bi-adjacency matrix $\bB^{\text{syn}}$ as $\bb_1, \cdots, \bb_{n_1^{\text{syn}}}$. We introduce the constraint $\bm{1}^T \bb_i = 1$ for all $i \in [n_1^{\text{syn}}]$. After applying the unbiased sampling algorithm, this constraint will translate into the condition that each record in the first synthetic table has exactly one relationship with a record in the second table. Finally, we denote the concatenation of $\bb_1, \cdots, \bb_{n_1^{\text{syn}}}$ by $\bb$ (i.e., $\bb$ is the vectorization of $\bB^{\text{syn}}$).
We solve the following relaxed optimization:
\begin{equation}
\label{eq::relaxed_opt_one-to-many}
\begin{aligned}
    \min_{\bb \in \Omega}~f(\bb) \quad \text{where }&\ f(\bb) \defined \left\| \frac{1}{m^{\text{syn}}}\hat{\bQ} \bb - \hat{\ba} \right\|_2^2\\
    \text{and }&\ \Omega \defined \{\bb \mid 0 \leq \bb \leq 1 \text{ and } \ones^T  \bb_{i} = 1 \ \forall i \in [ n_1^{\text{syn}}]\}.
\end{aligned}
\end{equation}
Note that the constraint set $\Omega$ can be reformulated for each individual vector $\bb_i$:
\begin{align*}
    \Omega_i \defined \{\bb_i \mid 0 \leq \bb_i \leq 1, \ones^T  \bb_{i} = 1\}. 
\end{align*}
Therefore, we can still apply the projected gradient descent algorithm to solve the relaxed optimization problem in \eqref{eq::relaxed_opt_one-to-many}. In each iteration, after computing the gradient \eqref{eq::relaxed_grad} based on the last iterate, we update the solution in the direction opposite to the gradient. Instead of projecting the entire vector $\bb$ onto $\Omega$, we project each sub-vector $\bb_i$ onto its respective constraint set $\Omega_i$. This can be achieved by applying the same closed-form expression from Theorem~\ref{thm::proj_cf} to each $\bb_i$. The projection step maintains the same time complexity but offers the added advantage of allowing parallel projection of the sub-vectors onto their respective constraint sets. Lastly, since the objective function remains unchanged, the convergence guarantee and learning rate are preserved.

Our projected gradient descent algorithm ensures that each row of $\tilde{\bB}^{\text{syn}}$ is a valid probability vector. Hence, we can directly sample from the categorical distribution parameterized by each row. We introduce Algorithm~\ref{alg::categoricalround} that samples from $\tilde{\bB}^{\text{syn}}$ while ensuring the one-to-many relationship. This approach provides an unbiased sample and has the added benefit of enabling parallel sampling for each record in the first table.

\begin{algorithm}[httb]
\begingroup
\small
\caption{Categorical rounding for one-to-many relationship.}
\label{alg::categoricalround}

\begin{algorithmic}[*]

\State {\bfseries Input:} relaxed bi-adjacency matrix $\tilde{\bB}^{\text{syn}} \in [0,1]^{n^{\text{syn}}_1} \times [0,1]^{n^{\text{syn}}_2}$

\State{Assert $\operatorname{sum}(\tilde{\bB}^{\text{syn}}_{i,:}) == 1$ for $i\in \{1,\cdots,n_1^{\text{syn}}\}$}

\State {$\bB^{\text{syn}} = \mathsf{zeros}(n_1^{\text{syn}}, n_2^{\text{syn}})$}

\For{$i = 1, \cdots, n_1^{\text{syn}}$}

\State Draw $k$ from $\{1, \cdots, n_2^{\text{syn}}\}$ according to the probability distribution $\tilde{\bB}^{\text{syn}}_{i,:}$

\State{$B_{i, k}^{\text{syn}} = 1$}

\EndFor{}

\State {\bfseries Output:} unweighted bi-adjacency matrix $\bB^{\text{syn}}$

\end{algorithmic}
\endgroup
\end{algorithm}

In short, generating synthetic data becomes significantly easier when the original relational database only contains one-to-many relationships. In our case, both the projected gradient descent algorithm and the unbiased sampling algorithm are simplified, enabling more efficient execution through parallel computation.

\section{Omitted Proofs}
\label{apx::proofs}
We provide omitted proofs in this section.

\subsection{Proof of Lemma~\ref{lem::qv}}
\label{apx::qv}
\begin{proof}
Recall from Section~\ref{sec::dp} that for any $k$-way cross-table query $\bq_{(\mathcal{S}_1, \mathcal{S}_2), (\by_1, \by_2)}$, its average over a relational database $(\mathcal{D}_1,\mathcal{D}_2,\bB)$ can be computed as 
\begin{align}
\label{eq::qv_1}
    \bq_{(\mathcal{S}_1, \mathcal{S}_2), (\by_1, \by_2)}(\mathcal{B}) = \frac{1}{\ones^T \cdot \operatorname{vec}(\bB)} \sum_{i=1}^{n_1}\sum_{j=1}^{n_2} \bq_{(\mathcal{S}_1, \mathcal{S}_2), (\by_1, \by_2)}(\bx_i,\bx_j) B_{i,j}
\end{align}
where $\bx_i\in \mathcal{D}_1$ and $\bx_j \in \mathcal{D}_2$. We define $\bQ_{(\mathcal{S}_1, \mathcal{S}_2), (\by_1, \by_2)} \defined \ones_{(\mathcal{S}_1, \by_1)} \cdot \ones^T_{(\mathcal{S}_2, \by_2)} \in \Reals^{n_1\times n_2}$ with its $i$-th row and $j$-th column denoted by $\left[\bQ_{(\mathcal{S}_1, \mathcal{S}_2), (\by_1, \by_2)}\right]_{i,j}$. By the definition of $\ones_{(\mathcal{S}_1, \by_1)}$ and $\ones_{(\mathcal{S}_2, \by_2)}$, we have $\left[\bQ_{(\mathcal{S}_1, \mathcal{S}_2), (\by_1, \by_2)}\right]_{i,j} = \bq_{(\mathcal{S}_1, \mathcal{S}_2), (\by_1, \by_2)}(\bx_i,\bx_j)$. Therefore, 
\begin{align*}
    \operatorname{vec}\left(\bQ_{(\mathcal{S}_1, \mathcal{S}_2), (\by_1, \by_2)}\right)^T \cdot \operatorname{vec}(\bB)
    &= \sum_{i=1}^{n_1}\sum_{j=1}^{n_2} [\bQ_{(\mathcal{S}_1, \mathcal{S}_2), (\by_1, \by_2)}]_{i,j} B_{i,j}\\
    &= \sum_{i=1}^{n_1}\sum_{j=1}^{n_2} \bq_{(\mathcal{S}_1, \mathcal{S}_2), (\by_1, \by_2)}(\bx_i,\bx_j) B_{i,j}.
\end{align*}
Combined this with \eqref{eq::qv_1} yields the desired result.
\end{proof}

\subsection{Proof of Theorem~\ref{thm::DP_guarantee}}
\label{apx::dpproof}
We recall Theorem~\ref{thm::DP_guarantee} below and provide additional details along with the proof.
\begin{theorem}
\label{thm::DP_guarantee_append}
Assume that each record in the original relational database has at most $d_{\max}$ relationships with records in the other table. We apply any DP mechanism with privacy budgets $(\epsilon_1, \delta_1)$ and $(\epsilon_2, \delta_2)$ to generate individual synthetic tables, respectively. Then we apply Algorithm
~\ref{alg::adapt_alg_l2} with privacy budget $(\epsilon_{\text{rel}}, \delta_{\text{rel}})$ to build their relationship. The synthetic relational database satisfies $(\epsilon_1+\epsilon_2+\epsilon_{\text{rel}}, \delta_1+\delta_2+\delta_{\text{rel}})$-DP, where $\epsilon_{\text{rel}}(\delta_{\text{rel}}) \leq \rho_{\text{rel}} + 2\sqrt{\rho_{\text{rel}}\log(1/\delta_{\text{rel}})}$. This condition can be achieved by choosing $\rho_{\text{rel}} = \left(\sqrt{\log(1/\delta_{\text{rel}}) + \epsilon_{\text{rel}}} - \sqrt{\log(1/\delta_{\text{rel}})}\right)^2$.
\end{theorem}
\begin{proof}
At each iteration, Algorithm
~\ref{alg::adapt_alg_l2} makes $K$ call to the exponential mechanism and $K$ call to the Gaussian mechanism. The exponential mechanism uses the score function:
\begin{align}
    \mathsf{score}(w) = \frac{1}{2} \|P_{w}(\mathcal{B}) - P_{w}(\mathcal{B}^{\text{syn}})\|_1.
\end{align}
Recall that we consider two relational databases $\mathcal{B}$ and $\mathcal{B}'$ adjacent if $\mathcal{B}'$ can be obtained from $\mathcal{B}$ by selecting a table, modifying the values of a single row within this table, and changing all relationship associated with this row. We focus on the setting of bounded DP so both $\mathcal{B}$ and $\mathcal{B}'$ have $m$ relationships. Additionally, we assume each record can have at most $d_{\max}$ relationships with records in the other table. Hence, for any workload $w$, the total variance distance between the empirical distributions of $\mathcal{B}$ and $\mathcal{B}'$ on $w$ has an upper bound: 
\begin{align}
    \frac{1}{2}\|P_{w}(\mathcal{B}) - P_{w}(\mathcal{B}')\|_1
    \leq \frac{d_{\max}}{m}. 
\end{align}
By the triangle inequality, we can upper bounded the sensitivity of $\mathsf{score}$ by $\frac{d_{\max}}{m}$. Similarly, for each workload $w$, we can upper bound the $L_2$ sensitivity of each empirical distribution $P_{w}(\mathcal{B})$ by $\frac{\sqrt{2}d_{\max}}{m}$. By Lemmas~\ref{lem::Gaussian} and ~\ref{lem::exponential}, at each iteration, the exponential and Gaussian mechanisms satisfy $\frac{K}{2}\alpha \cdot \epsilon_0^2$-zCDP and $\frac{K}{2} (1-\alpha)\cdot \epsilon_0^2$-zCDP, respectively. The composition theorem in Lemma~\ref{lem::comp-zcdp} ensures that Algorithm
~\ref{alg::adapt_alg_l2} satisfies 
$\frac{K T(\alpha + (1-\alpha)) }{2}\epsilon_0^2 $-zCDP. Recall that we chose $\epsilon_0 = \sqrt{\frac{2\rho_{\text{rel}}}{KT}}$. Hence, Algorithm~\ref{alg::adapt_alg_l2} satisfies $\rho_{\text{rel}}$-zCDP. As a result, Lemma~\ref{lem::approx-zcdp} yields that it also satisfies $(\rho_{\text{rel}}+2 \sqrt{\rho_{\text{rel}} \log (1 / \delta)}, \delta)$-DP. In particular, if we choose $\rho_{\text{rel}} = \left(\sqrt{\log(1/\delta_{\text{rel}}) + \epsilon_{\text{rel}}} - \sqrt{\log(1/\delta_{\text{rel}})}\right)^2$, then Algorithm~\ref{alg::adapt_alg_l2} will satisfy $(\epsilon_{\text{rel}}, \delta_{\text{rel}})$-DP.
\end{proof}

\subsection{Proof of Theorem~\ref{thm::proj_cf}}
\label{apx::proj_cf}
\begin{proof}
The Euclidean projection of a vector $\bb$ onto the set $\Omega$ can be written as solving the following optimization:
\begin{align*}
    \min_{\bb_p}~ \| \bb_p - \bb \|_2^2 \quad \sto~0 \leq \bb_p \leq 1, \ones^T \bb_p = m^{\text{syn}}.
\end{align*}
We denote $\bm{\delta} \triangleq \bb_p - \bb $. Then we can rewrite the above optimization as
\begin{align*}
    \min_{\bm{\delta}}&~\| \bm{\delta} \|_2^2 \\
    \sto&~- \bb \leq \bm{\delta} \leq \ones - \bb\\
        &~\ones^T \bm{\delta} = m^{\text{syn}} - \ones^T \bb.
\end{align*}
Theorem 2.1 from \citet{pardalos1990algorithm} implies that $\bm{\delta}^*$ is the global optimal solution if and only if there exists a $y \in \Reals$ such that $\forall i \in [N]$
\begin{align}
\label{eq::proj_delta}
     \delta_i^*(y) = 
    \begin{cases}
        -b_i \quad  &\text{if } y < -b_i,\\
        y \quad  &\text{if } -b_i \leq y \leq 1-b_i,\\
        1-b_i \quad  &\text{if } y > 1 -b_i.
    \end{cases}
\end{align}
and $\sum_{i=1}^N \delta_i^*(y) = m^{\text{syn}} - \ones^T \bb$. Note that each $\delta_i^*(y)$ is a piecewise linear, monotone non-decreasing function of $y$. As a result, the function $\sum_{i=1}^N \delta_i^*(y)$ is also a piecewise linear, monotone non-decreasing function. This property allows us to use binary search to efficiently find a $y$ within the interval $[-\max_{i\in[N]}\{b_i\} , 1-\min_{i\in[N]}\{b_i\}]$ that satisfies $\sum_{i=1}^N \delta_i^*(y) = m^{\text{syn}} - \ones^T \bb$. Once we find such a value of $y$, we can substitute it into \eqref{eq::proj_delta} and use $\bb_p = \bm{\delta} + \bb$ to determine the projection vector $\bb_p$.
\end{proof}

\subsection{Proof of Theorem~\ref{thm::convergence}}
\label{apx::convergence}

\paragraph{Proof of convergence rate.} We begin by revisiting a classical convergence result for solving convex optimization problems using the projected gradient descent algorithm \citep[Theorem 3.7 in][]{bubeck2015convex}.
\begin{lemma}
\label{lem::pgdconvergence}     
For a convex function $f$ and a convex set $\Omega$, consider the problem of solving $\min_{\bb \in \Omega} f(\bb)$ using projected gradient descent, with the updating rule given by $\bb_{t+1} = \Pi_\Omega \left( \bb_t - \eta \nabla f(\bb_t) \right)$. Assume that $f(\cdot)$ is $L$-smooth: $\| \nabla f(\bx) - \nabla f(\by) \| \leq L \cdot \| \bx - \by \|$. If we set the step size $\eta = \frac{1}{L}$, the following bound holds:
\begin{align*}
    f(\bb_T) - f(\bb^{*}) \leq \frac{3L \| \bb_0 - \bb^* \|_2^2 + f(\bb_0) - f(\bb^{*}) }{T},
\end{align*}
where $\bb^{*} \in \argmin_{\bb \in \Omega} f(\bb)$ is an optimal solution to the optimization problem.
\end{lemma}

Now we prove the convergence rate in Theorem~\ref{thm::convergence} using the above lemma.
\begin{proof}
We first show that our objective function $f(\bb) = \|\frac{1}{m^{\text{syn}}} \hat{\bQ} \bb - \hat{\ba} \|_2^2$ is a $L$-smooth function with $L = 2 \cdot \left(\frac{\sigma_{\text{max}}(\hat{\bQ})}{m^{\text{syn}}}\right)^2$ where $\sigma_{\text{max}}(\hat{\bQ})$ is the largest singular value of $\hat{\bQ}$. For any $\bx, \by$,
\begin{align*}
    \left\| \nabla f(\bx) - \nabla f(\by) \right\|_2
    &= \left\|  \frac{2}{m^{\text{syn}}} \hat{\bQ}^T \left(\frac{1}{m^{\text{syn}}} \hat{\bQ} \bx - \hat{\ba}\right) - \frac{2}{m^{\text{syn}}} \hat{\bQ}^T \left(\frac{1}{m^{\text{syn}}} \hat{\bQ} \by - \hat{\ba}\right) \right\|_2 \\
    &=  \frac{2}{(m^{\text{syn}})^2} \cdot \left\| \hat{\bQ}^T \hat{\bQ} (\bx - \by) \right\|_2 \\
    &\leq \frac{2}{(m^{\text{syn}})^2} \cdot \left\|\hat{\bQ}^T \hat{\bQ}\right\|_2 \cdot \| \bx - \by \|_2 \\
    &= 2 \left(\frac{\sigma_{\text{max}}(\hat{\bQ})}{m^{\text{syn}}}\right)^2 \cdot \| \bx - \by \|_2.
\end{align*}
Now by applying Lemma~\ref{lem::pgdconvergence}, after $T$ iteration of projected gradient descent, we obtain the desired inequality:
\begin{align*}
     f(\bb_T) - f(\bb^{*}) \leq \frac{3L \| \bb_0 - \bb^{*} \|_2^2 + f(\bb_0) - f(\bb^{*}) }{T}.
\end{align*}
%
\end{proof}

\paragraph{Runtime analysis.} Let the query matrix $\hat{\bQ} \in \Reals^{d \times N}$, where $d$ denotes the number of cross-table $k$-way marginal queries selected by the exponential mechanism, and $N$ represents the dimension of the bi-adjacency matrix for the synthetic relational database. Our algorithm begins by running the power iteration method to compute $\sigma_{\text{max}}(\hat{\bQ})$, which is used to determine the step size. In each iteration, the power iteration performs the operation $\hat{\bQ}^T (\hat{\bQ} \bv)$, followed by normalizing the resulting vector. The matrix-vector multiplications require $O(d \cdot N)$ operations, while normalization takes $O(N)$ operations. Thus, the overall computational complexity of this step is $O(T_{\text{power}} \cdot d \cdot N)$, where $T_{\text{power}}$ denotes the number of power iterations.

We run the projected gradient descent algorithm for $T$ iterations. In each iteration, the gradient of the objective function is computed as $\nabla f(\bb_t) = \frac{2}{m^{\text{syn}}} \hat{\bQ}^T (\frac{1}{m^{\text{syn}}} \hat{\bQ} \bb_t - \hat{\ba})$. The matrix-vector multiplication involved in this computation has a complexity of $O(d \cdot N)$. Afterward, we project the solution onto the constraint set $\Omega$. Using the closed-form expression from Theorem~\ref{thm::proj_cf}, the projection of $\bb_t - \eta \nabla f(\bb_t)$ onto $\Omega$ has $O(N)$ complexity. Therefore, the total computational complexity is $O(T \cdot d \cdot N)$. We omit multiplicative constants $T_{\text{power}}$, $T$, and $d$ in our final expression as they are relatively way much smaller than $N$.

\subsection{Proof of Theorem~\ref{thm::ubs}}
\label{append::samplingproof}
\paragraph{Proof of unbiased sampling.} We prove it via mathematical induction. For the two base cases $m=0$ and $m=1$, $\mathsf{UBS}(\bx,m)$ always returns an unbiased sample. Now assume that for $m - 1 \geq 1$, $\mathsf{UBS}(\bx,m-1)$ returns an unbiased sample for any vector $\bx$. We will prove that the same holds for $\mathsf{UBS}(\bx,m)$.

In the first step of our algorithm (merging), the indices of $\bx$ are partitioned into groups $\mathcal{G}[0], \cdots, \mathcal{G}[L-1]$. We prove that $2 \leq m \leq L \leq 2 m - 1$. From \eqref{eq::defn_sum_Gj}, we have
\begin{align*}
    m = \ones^T \bx 
    = \sum_{j=0}^{L-1}\mathsf{sum}(\bx\mid \mathcal{G}[j])
    \leq L,
\end{align*}
where the last step is because $\mathsf{sum}(\bx\mid \mathcal{G}[j]) \leq 1$. Therefore, we have $2 \leq m \leq L$. Note that 
\begin{align}
\label{eq::merge_prob}
    \mathsf{sum}(\bx\mid \mathcal{G}[j]) + \mathsf{sum}(\bx\mid \mathcal{G}[j+1]) > 1  \quad \text{for } j = 0,\cdots,L-2
\end{align}
since otherwise $\mathcal{G}[j]$ and $\mathcal{G}[j+1]$ would have been merged in the previous step. \\
\textbf{Case 1.} If $L\geq 2$ is even, then
\begin{align*}
    m = \sum_{j=0}^{L-1} \mathsf{sum}(\bx\mid \mathcal{G}[j]) 
    = \sum_{j=0}^{L/2-1} (\mathsf{sum}(\bx\mid \mathcal{G}[2j]) +  \mathcal{G}[2j+1])
    > L/2.
\end{align*}
Thus, $2m > L$, and since $L$ is an integer, we have $2m-1 \geq L$.\\
\textbf{Case 2.} If $L \geq 2$ is odd, 
\begin{align*}
    m 
    &= \sum_{j=0}^{L-1} \mathsf{sum}(\bx\mid \mathcal{G}[j]) 
    = \sum_{j=0}^{(L-1)/2-1} (\mathsf{sum}(\bx\mid \mathcal{G}[2j]) +  \mathcal{G}[2j+1]) + \mathsf{sum}(\bx\mid \mathcal{G}[L-1])\\
    &> (L-1)/2.
\end{align*}
This implies $2m + 1 > L$, and since $L$ is odd, we conclude that $2m-1 \geq L$.\\
According to Algorithm~\ref{alg::rec_ubs}, $\mathsf{UBS}(\bx,m)$ calls $\mathsf{UBS}(\bp', L - m)$ to sample $L-m$ groups to exclude where $p'_j = 1 - \mathsf{sum}(\bx\mid\mathcal{G}[j])$. Since we just proved that $0\leq L-m \leq m-1$, by our assumption $\mathsf{UBS}(\bp', L - m)$ returns an unbiased sample:
\begin{align*}
    \mathbb{P}(\text{group $\mathcal{G}[j]$ is selected})
    = 1 - \mathbb{P}(\text{group $\mathcal{G}[j]$ is excluded})
    = 1 - p'_j 
    = \mathsf{sum}(\bx\mid\mathcal{G}[j]). 
\end{align*}
Finally, by our sampling strategy, for the index $i\in \mathcal{G}[j]$
\begin{align*}
    \mathbb{P}(\text{$i$ is sampled})
    &= \mathbb{P}(\text{group } \mathcal{G}[j] \text{ is selected})  * \mathbb{P}(\text{$i$ is sampled} \mid \text{group $\mathcal{G}[j]$ is selected}) \\
    &= \mathsf{sum}(\bx\mid \mathcal{G}[j]) * \frac{x_i}{\groupsum{j}}
    = x_i.
\end{align*}
This condition holds for any index $i$. Thus, $\mathsf{UBS}(\bx, m)$ returns an unbiased sample, completing the inductive step. As a corollary, our algorithm will terminate since each recursive call to $\mathsf{UBS}$ reduces the value of its second argument by at least one.

\paragraph{Runtime analysis.} We analyze the runtime of Algorithm~\ref{alg::rec_ubs}. Since the vector $\bb^{\text{syn}}$ has dimension $N$, the first time $\mathsf{UBS}$ merges the indices of $\bb^{\text{syn}}$ into groups, the runtime is $O(N)$. This merging process results in $L$ groups, and we have already proven that $0\leq L-m \leq m-1$. Therefore, when the recursive algorithm $\mathsf{UBS}$ is called for the second time, its input vector $\bx$ has a dimension smaller than $m$. Below, we analyze its runtime. The key idea in our analysis is that the number of groups after merging decreases exponentially by a factor of at least $2/3$ in each iteration.

For any $n$, let $T(n)$ denote the runtime of $\mathsf{UBS}$ on a vector $\bx$ of size $n$, which is the output of the recursive algorithm in the previous step. We omit the dependence on the second argument of $\mathsf{UBS}$, as it is always smaller than $n$.

Since $\bx$ is the output of the previous recursion, it represents a vector of complement probabilities with dimension $n$. As a result, $x_j + x_{j+1} < 1$; otherwise, the $j$-th and $(j+1)$-th groups would have already been merged in the previous iteration (similar to the reasoning in \eqref{eq::merge_prob}). In the current iteration, when we partition the indices of $\bx$ into disjoint groups, each group, except the last one, will contain at least two indices. Thus, the number of groups in the current iteration, $n'$, satisfies $n' \leq 1 + \frac{n - 1}{2} \leq \frac{2}{3}n$ for $n\geq 3$. As a result, when the current iteration calls the function $\mathsf{UBS}$ itself, the length of its input vector will be at most $\frac{2}{3}n$.

On the other hand, the merging step can be completed in a single scan of the vector $\bx$, which has dimension $n$, resulting in a runtime of $O(n)$. Hence,
\begin{align*}
T(n) 
= T\left(\frac{2}{3}n\right) + O(n).
\end{align*}
By the master theorem \citep{cormen2022introduction}, the overall runtime is $T(n) = O(n)$ for any $n$. Putting everything together, the runtime of Algorithm~\ref{alg::rec_ubs} is $O(N) + T(m) = O(N)$, since $T(m) = O(m)$ and $m \leq N$.

\subsection{Proof of Theorem~\ref{thm::concentration}}
\label{apx::NA}

Before presenting the proof of Theorem~\ref{thm::concentration}, we first recall the definition of negatively associated (NA) random variables, explaining why this property is crucial for deriving Hoeffding's inequality in the context of dependent random variables. The key step in proving Theorem~\ref{thm::concentration} is to establish that the outputs from Algorithm~\ref{alg::rec_ubs} are NA random variables. Since Algorithm~\ref{alg::rec_ubs} is a recursive algorithm that alternates between complementing and merging, we next introduce two useful lemmas.

Lemma~\ref{lem::NAinv} shows that if a set of binary random variables $(X_1, \cdots, X_n)$ is NA, then their complements $(1 - X_1, \cdots, 1 - X_n)$ also satisfy the NA property. This result ensures that the complement step in Algorithm~\ref{alg::rec_ubs} preserves the NA property. Lemma~\ref{lem::NAred} shows that if the random variables satisfy the NA property after a merging step in Algorithm~\ref{alg::rec_ubs}, then they maintain the NA property prior to this merging. Together, these lemmas reduce the problem to proving that the random variables in the base case, $m=1$, are NA.

\begin{definition}
Random variables $X_1, X_2, \cdots, X_N$ are said to satisfy \emph{negatively associated} (NA) if for any $\mathcal{A}_1, \mathcal{A}_2$ which are disjoint subsets of $[N]$, we have
\begin{equation}
    \operatorname{Cov}\left( f_1(X_i, i \in \mathcal{A}_1), f_2(X_j, j \in \mathcal{A}_2) \right) \leq 0,
\end{equation}
whenever \(f_1\) and \(f_2\) are both non-increasing or both non-decreasing. This inequality is also equivalent to
\begin{equation}
    \mathbb{E}[f_1(X_i, i \in \mathcal{A}_1) \cdot f_2(X_j, j \in \mathcal{A}_2)] \leq \mathbb{E}[f_1(X_i, i \in \mathcal{A}_1)] \cdot \mathbb{E}[f_2(X_j, j \in \mathcal{A}_2)].
\end{equation}
\end{definition}

\begin{lemma}
\label{lem::NAHoeffding}
\citep{dubhashi1998NA} Let \(X_1, X_2, \dots, X_N\) be NA binary random variables. The Hoeffding's inequality is applicable to their sum:
\begin{align*}
    \mathbb{P} \left( \left|\sum_i X_i - \EE{\sum_i X_i}\right| \geq t \right) \leq 2 \exp\left(\frac{-2t^2}{N}\right).
\end{align*}
\end{lemma}
\begin{lemma}
    \label{lem::NAsubset} \citep{Joagdev1983NegativeAO} If \(X_1, X_2, \dots, X_N\) are NA random variables, for any subset of indices $\mathcal{I} \subseteq [N]$, the set of random variables $\{X_i\}_{i \in I}$ also satisfies NA.
\end{lemma}
Combining Lemma~\ref{lem::NAHoeffding} and \ref{lem::NAsubset} leads to the following result.
\begin{lemma}
\label{lem::NAHoef} 
Let \(X_1, X_2, \dots, X_N\) be NA binary random variables. For any subset of indices $\mathcal{I} \subseteq [N]$, the sum of random variables in $\mathcal{I}$ satisfies Hoeffding's inequality:
\begin{align*}
    \mathbb{P}\left( \left|\sum_{i \in \mathcal{I}} X_i - \EE{\sum_{i \in \mathcal{I}} X_i}\right| \geq t \right) \leq 2 \exp\left(\frac{-2t^2}{N}\right).
\end{align*}
\end{lemma}
\begin{proof}
Lemma~\ref{lem::NAsubset} implies that $\{X_i\}_{i \in \mathcal{I}}$ are NA. Therefore, by applying Lemma~\ref{lem::NAHoeffding} on them, we have
\begin{align*}
     \mathbb{P}\left( \left|\sum_{i \in \mathcal{I}} X_i - \EE{\sum_{i \in \mathcal{I}} X_i}\right| \geq t \right) \leq 2 \exp\left(\frac{-2t^2}{|\mathcal{I}|}\right).
\end{align*}
Since $\mathcal{I} \subseteq [N]$, we have $|\mathcal{I}| \leq N$ and, hence, 
\begin{align*}
    \mathbb{P}\left( \left|\sum_{i \in \mathcal{I}} X_i - \EE{\sum_{i \in \mathcal{I}} X_i}\right| \geq t \right) \leq 2 \exp\left(\frac{-2t^2}{|\mathcal{I}|}\right) \leq  2 \exp\left(\frac{-2t^2}{N}\right),
\end{align*}
which completes the proof.
\end{proof}

The next result shows that the complement step in Algorithm~\ref{alg::rec_ubs} preserves the NA property.
\begin{lemma}
\label{lem::NAinv} 
If \(X_1, X_2, \cdots, X_N\) are NA random variables, then  \(1 - X_1, 1 - X_2, \cdots, 1 - X_N\) also satisfy NA.
\end{lemma}

\begin{proof}
For any non-decreasing functions $f_1$ and $f_2$ and two disjoint subsets $\mathcal{A}_1, \mathcal{A}_2 \subseteq [N]$, we can define $g_1$ and $g_2$ as follows:
\[
g_{i}(X_i, i \in \mathcal{A}_i) \defined f_{i}(1 - X_i, i \in \mathcal{A}_i) \quad \text{for } i \in \{1,2\}.
\]
By definition, $g_1$ and $g_2$ are non-increasing functions. Since $X_1,\cdots,X_N$ are NA, we have
\[
 \operatorname{Cov}\left( f_1(1-X_i, i \in \mathcal{A}_1), f_2(1-X_j, j \in \mathcal{A}_2) \right) =  \operatorname{Cov}\left( g_1(X_i, i \in \mathcal{A}_1), g_2(X_j, j \in \mathcal{A}_2) \right) \leq 0. 
\]
The proof for non-increasing $f_1$ and $f_2$ can be derived in a similar manner. 
\end{proof}

\begin{lemma}
\label{lem::NAred} 
Let $X_1, X_2, \cdots, X_N$ be binary NA random variables, and let $Z$ be a random variable, independent with $X_1,\cdots, X_n$, taking values in $[n]$ with probability distribution:
\begin{align*}
\mathbb{P} (Z = i) = \frac{p_i}{\sum_{j \in [n]} p_j} \quad \forall i \in [n].
\end{align*}
We define $Y_1, \cdots, Y_n $ as binary random variables where $Y_i = 1$ if and only if $Z = i$ and $X_N = 1$. Then $X_1, X_2, \cdots, X_{N-1}, Y_1, \cdots, Y_n$ are also NA.
\end{lemma}

\begin{proof}
Let $f_1$ and $f_2$ be non-decreasing functions. Let \(\mathcal{A}_1, \mathcal{A}_2\) be disjoint subsets of $[N-1]$ and \(\mathcal{B}_1, \mathcal{B}_2\) be disjoint subsets of $[n]$. We aim to prove that
\[
\operatorname{Cov}\left( f_1(X_{i_\mathcal{A}}, Y_{i_\mathcal{B}}, i_{\mathcal{A}} \in \mathcal{A}_1, i_{\mathcal{B}} \in \mathcal{B}_1), f_2(X_{j_{\mathcal{A}}}, Y_{j_{\mathcal{B}}}, j \in \mathcal{A}_2, {j_{\mathcal{B}}} 
\in \mathcal{B}_2) \right) \leq 0.
\]
Using the law of total covariance and conditioning on $Z$, we have
\begin{equation}
\label{eq::cond_cov}
\operatorname{Cov}\left( f_1(\cdot ), f_2(\cdot) \right) = \mathbb{E}[ \operatorname{Cov}\left( f_1(\cdot ), f_2(\cdot) \mid Z  \right)] + \operatorname{Cov}\left( \mathbb{E}[f_1(\cdot ) \mid Z ], \mathbb{E}[f_2(\cdot ) \mid Z ] \right). 
\end{equation}
Here we simplify notation by denoting $f_1(\cdot ) \defined f_1(X_{i_\mathcal{A}}, Y_{i_\mathcal{B}}, i_{\mathcal{A}} \in \mathcal{A}_1, i_{\mathcal{B}} \in \mathcal{B}_1)$, with $f_2(\cdot)$ defined analogously. 
Note that $Y_1, \cdots, Y_n$ can be fully determined by $X_N$ and $Z$ since we can write:
\begin{align*}
    Y_Z = X_N\quad \text{and} \quad Y_{i} = 0 \text{ for } i\neq Z.
\end{align*}
Without loss of generality, assume that $z \in \mathcal{B}_1$. We define $\mathcal{A}'_1 = \mathcal{A}_1 \cup \{N\}$. Conditioning on $Z = z$, we define
\begin{align*}
    f^z_1 (X_i, i \in \mathcal{A}'_1) &= f_1(X_{i_{\mathcal{A}}}, Y_{i_{\mathcal{B}}}, i_{\mathcal{A}} \in \mathcal{A}_1, i_{\mathcal{B}} \in \mathcal{B}_1). \\
    f^z_2 (X_i, i \in \mathcal{A}_2) &= f_2(X_{j_{\mathcal{A}}}, Y_{j_{\mathcal{B}}}, j_{\mathcal{A}} \in \mathcal{A}_2, j_{\mathcal{B}} \in \mathcal{B}_2).
\end{align*}
These two functions are still non-decreasing since they can be obtained by fixing $n-1$ variables (i.e., $Y_i$ for $i\in[n] \backslash \{z\}$) to zero and changing the variable name from $Y_z$ to $X_N$. Since \(X_1, \cdots, X_N \) are NA random variables, we have
\begin{align*}
\operatorname{Cov} \left(f^z_1 (X_i, i \in \mathcal{A}'_1),  f^z_2 (X_i, i \in \mathcal{A}_2) \right) \leq 0,\quad \forall z \in [n].
\end{align*}
Averaging this inequality over the distribution of $Z$ leads to
\begin{align*}
\mathbb{E}[ \operatorname{Cov}\left( f_1(\cdot ), f_2(\cdot) \mid Z  \right)] \leq 0.
\end{align*}
Next, we prove that the second term on the right-hand-side of \eqref{eq::cond_cov} is also non-positive. 
We define 
\begin{align*}
g_i(Z) \defined \mathbb{E}[f_i(\cdot ) \mid Z ] - \mathbb{E}[f_i(\cdot ) \mid \forall j \in \mathcal{B}_i: Y_j = 0].
\end{align*}
Since covariance is linear shift-invariant, we have
\begin{align}
\operatorname{Cov}\left( \mathbb{E}[f_1(\cdot ) \mid Z ], \mathbb{E}[f_2(\cdot ) \mid Z ] \right)
&= \operatorname{Cov}\left( g_1(Z), g_2(Z)\right)\nonumber\\
&= \EE{g_1(Z) g_2(Z)} - \EE{g_1(Z)} \EE{g_2(Z)}. \label{eq::cov_deco}
\end{align}
If $z\notin \mathcal{B}_1$, then $Y_j = 0$ for $\forall j \in \mathcal{B}_i$. In this case, $g_1(z) = 0$ by its definition. Similarly, if $z\notin \mathcal{B}_2$, then $g_2(z) = 0$. Since $\mathcal{B}_1$ and $\mathcal{B}_2$ are disjoint, at least one of the events $Z \notin \mathcal{B}_1$ or $Z \notin \mathcal{B}_2$ occurs. Therefore,
\begin{align*} 
g_1(z) g_2(z)= 0,\quad \forall z \in [n].
\end{align*}
Taking expectation over $Z$, we have
\begin{align} 
\label{eq::cov_1}
 \EE{g_1(Z) g_2(Z)} = 0.
\end{align}
Since $f_i$ is a non-decreasing function, $g_i(z) \geq 0$ for $\forall z\in [n]$. Therefore, $\EE{g_i(Z)} \geq 0$, yielding
\begin{align}
\label{eq::cov_2}
    \EE{g_1(Z)}\EE{g_2(Z)} \geq 0.
\end{align}
Substituting \eqref{eq::cov_1} and \eqref{eq::cov_2} into \eqref{eq::cov_deco} leads to the desired result. 
\end{proof}

We extend Lemma~\ref{lem::NAred} to analyze the merging step in Algorithm~\ref{alg::rec_ubs}. During this step, we partition $[N]$ into $L$ groups, $\mathcal{G}[1], \cdots, \mathcal{G}[L]$, where each group satisfies the conditions of Lemma~\ref{lem::NAred}. We sample an index within each group with probability proportional to its desired probability, normalized by the total probability of the subset. Lemma~\ref{lem::NAred} ensures that if the set of the random variables corresponding to the merged sets is NA, then unmerging one group and replacing its random variable with those of its elements still preserve the NA property. In Lemma~\ref{lem::NAredGen}, we present this result formally.
\begin{lemma}
\label{lem::NAredGen} 
Suppose $\mathcal{G}[1], \cdots, \mathcal{G}[L]$ are disjoint subsets that partition the set $[N]$. Let \(X_1, X_2, \cdots, X_L\) be binary NA random variables. Additionally, let \(Z_1, \cdots, Z_L \) be random variables, mutually independent and independent with \(X_1, X_2, \cdots, X_L\). Each $Z_i$ takes values in $\mathcal{G}[i]$ with probability:
\[
\mathbb{P} [Z_i = j] = \frac{p_{i,j}}{\sum_{j' \in G[i]} p_{i,j'}}\quad \forall j \in {G[i]}. 
\]
We define $\{Y_{i,j} \}_{i \in [L], j \in G[i]}$ as binary random variables where \(Y_{i,j} = 1 \) if and only if $Z_i = j$ and $X_i = 1$. Then  $\{Y_{i,j} \}_{i \in [L], j \in G[i]}$ are also NA.
\end{lemma}

\begin{proof}
We denote the number of subsets containing more than one element by $k \defined |\{i \mid |G[i]| \geq 2\}|$. We prove this lemma by induction on $k$. In the base case, where all groups have exactly one element (i.e., $k=0$), we have $\mathbb{P} [Z_i = j] = 1$ for $j \in {G[i]}$. Then, by definition, $Y_{i,j}  = X_i$ for $j \in {G[i]}$. Consequently, $\{Y_{i,j} \}_{i \in [L], j \in G[i]}$ is equivalent to \(\{X_i\}_{i \in [L]}\) and is NA.

Assume that the lemma holds for $k$. Next, we prove that it holds for the case where $k+1$ subsets have more than one element. Without loss of generality, suppose that $G[L] = \{L_1, \cdots,L_s \}$ with $s\geq 2$. By Lemma~\ref{lem::NAred}, replacing $X_L$ with $\{Y_{L,j} \}_{j \in {G[L]}}$ in the set $\{X_1, X_2, \cdots, X_L\}$ maintains the NA property. This new set of random variables corresponds to the following disjoint subsets that partition $[N]$:
\begin{align*}
\mathcal{G}[1],\cdots,\mathcal{G}[L-1], \{L_1\},\cdots, \{L_s\}.
\end{align*}
Among them, there are only $k$ subsets having more than one element. The induction hypothesis implies that unpacking the remaining subsets with more than one element will also preserve the NA property. 
\end{proof}

Now we provide a proof of Theorem~\ref{thm::concentration}. The key idea is to prove that the random variables in Algorithm~\ref{alg::rec_ubs} satisfy the NA property throughout the recursive sampling process. In the proof, we denote $\bb \defined \operatorname{vec}(\bB^{\text{syn}})$ and 
$\tilde{\bb} \defined \operatorname{vec}(\tilde{\bB}^{\text{syn}})$. 
%
%
\begin{proof}
We first prove that the function $\mathsf{UBS}(\bx,m)$ in Algorithm~\ref{alg::rec_ubs} outputs NA random variables. We use mathematical induction on $m$ to prove this. For the base case $m=1$, there is only one group, so sampling reduces to a categorical selection where one indicator variable equals $1$, and the rest are $0$. These random indicator variables satisfy NA \citep[see Section~3 (a) in][]{Joagdev1983NegativeAO}. Now assume that the NA property holds for $m$, and consider the case of $m + 1$. According to Algorithm~\ref{alg::rec_ubs}, we first apply a merging step followed by an inversion step. In the proof of Theorem~\ref{thm::ubs}, we established that after these steps, $m$ decreases, allowing us to apply the induction hypothesis. By Lemma~\ref{lem::NAredGen}, if the random variables are NA after the merging step, then the NA property holds for the variables prior to the merging step. Similarly, Lemma~\ref{lem::NAinv} ensures that the NA property is preserved through the inversion step. Therefore, by successively applying the merging and inversion steps, we conclude that the original set of random variables retains the NA property for $m+1$. Hence,  $\mathsf{UBS}(\bx,m)$ outputs NA random variables, and the vectorized bi-adjacency matrix $\bb$, produced by Algorithm~\ref{alg::rec_ubs}, satisfies the NA property. 

For any marginal query vector $\bq \in \mathcal{Q}$, by its definition in Lemma~\ref{lem::qv}, there exists a subset $\mathcal{I}_q \in [N]$ such that $\bq^T \bb = \sum_{i \in \mathcal{I}_q} b_i$. Additionally, recall that $\bb$ is an unbiased estimator of $\tilde{\bb}$ (Theorem~\ref{thm::ubs}). Hence, Lemma~\ref{lem::NAHoef} implies that
\begin{align*}
    \mathbb{P} \left( \left|\bq^T \bb - \bq^T \tilde{\bb} \right| \geq t \right) = \mathbb{P}\left( \left|\sum_{i \in \mathcal{I}_q} b_i - \EE{\sum_{i \in \mathcal{I}_q} b_i}\right| \geq t \right) \leq 2 \exp\left(\frac{-2t^2}{N}\right).
\end{align*}
This bound holds for any single query vector $\bq \in \mathcal{Q}$. Next, we apply the union bound to extend this result for all $\bq \in \mathcal{Q}$:
\begin{align*}
    \mathbb{P} \left( \exists \bq \in \mathcal{Q}: \left|\bq^T \bb - \bq^T \tilde{\bb} \right| \geq t \right) &\leq  \sum_{\bq \in \mathcal{Q}} \mathbb{P} \left( \left|\bq^T \bb - \bq^T \tilde{\bb} \right| \geq t \right)\\
    &\leq \sum_{\bq \in \mathcal{Q}} 2 \exp\left(\frac{-2t^2}{N}\right) \\
    &= 2 | \mathcal{Q} | \exp\left(\frac{-2t^2}{N}\right).
\end{align*}
Therefore, we have
\begin{align*}
    \mathbb{P} \left( \forall \bq \in \mathcal{Q}: \left|\frac{1}{m^{\text{syn}}}\bq^T \bb - \frac{1}{m^{\text{syn}}} \bq^T \tilde{\bb} \right| \leq \frac{t}{m^{\text{syn}}} \right) \geq 1 - 2 | \mathcal{Q} | \exp\left(\frac{-2t^2}{N}\right).
\end{align*}
In other words, with at least probability $1-\beta$, for $\forall \bq \in \mathcal{Q}$, we have
\begin{align*}
    \left| \frac{1}{m^{\text{syn}}} \bq^T \bb -  \frac{1}{m^{\text{syn}}}\bq^T \tilde{\bb}\right| 
    \leq \frac{1}{m^{\text{syn}}} \cdot \sqrt{\frac{N}{2} \cdot \log \frac{2 | \mathcal{Q}|}{\beta}},
\end{align*}
which completes the proof.
\end{proof}

\subsection{Proof of Theorem~\ref{thm::Projection_utility}}
\label{apx::utilityproof}
We provide a more rigorous statement of Theorem~\ref{thm::Projection_utility} along with its proof.
\begin{theorem}
\label{thm::proj_uti_append}
Let $\mathcal{D}^{\text{syn}}_1$ and $\mathcal{D}^{\text{syn}}_2$ be any individual synthetic tables. We stack all the $k$-way cross-table query vectors as $\hat{\bQ}$. Denote by $\ba$ the true answers to these queries computed from the original database. Suppose we use a privacy budget $(\epsilon_{\text{rel}}, \delta_{\text{rel}})$ to run a single iteration of Algorithm~\ref{alg::adapt_alg_l2} (without unbiased sampling), which outputs $\bB^{\text{syn}}$. Let $\bB^*$ denote the optimal weighted bi-adjacency matrix in minimizing the mean squared error when no privacy constraints are enforced. With probability at least $1 - \beta$, we have
\begin{align*}
    \left\| \frac{1}{m^\text{syn}} 
    \hat{\bQ} \operatorname{vec}(\bB^{\text{syn}}) - \ba \right\|_2^2
    \leq &8 \left\| \frac{1}{m^\text{syn}}\hat{\bQ}\operatorname{vec}(\bB^{*}) - \ba\right\|_2^2 \\
    &+ 8\sqrt{2}|\mathcal{W}_{\text{cross}, k}| \frac{d_{\text{max}}}{m\sqrt{\rho_{\text{rel}}}} \left( \operatorname{log}\frac{2 n_1^{\text{syn}} n_2^{\text{syn}}}{\beta}\right)^{\frac{1}{2}},
\end{align*}
where $\rho_{\text{rel}} = \left(\sqrt{\epsilon_{\text{rel}} + \operatorname{log} \frac{1}{\delta_{\text{rel}}}} -  \sqrt{\operatorname{log} \frac{1}{\delta_{\text{rel}}}}\right)^2$ and $m$ is the number of relationships in the original database.
\end{theorem}
We introduce some useful lemmas that will be used in the proof of Theorem~\ref{thm::proj_uti_append}.
\begin{lemma} (Contractive property of convex sets projection \cite{Schneider_2013})
\label{lem::contractive}
    Let $\mathcal{Y}$ be a non-empty closed convex subset of $\mathbb{R}^d$, then for any $\bx_1, \bx_2$ if we denote their projections by $\by_1, \by_2$ respectively, i.e.  
    $\by_i = \arg \min_{\by \in \mathcal{Y}} \|\bx_i - \by \|_2^2$, then we have 
    $$
    \|\by_1 - \by_2 \|_2^2 \leq  \|\bx_1 - \bx_2 \|_2^2.
    $$
\end{lemma}
\begin{proof}
Let $\bm{v} \defined \by_2-\by_2 \neq o$. The function $f$ defined by $f(t) \defined |\bx_1-(\by_1+t \bm{v})|^2$ for $t \in[0,1]$ has a minimum at $t=0$, hence $f^{\prime}(0) \geq 0$. This gives $\langle \bx_1-\by_1, \bm{v} \rangle \leq 0$. Similarly we obtain $\langle \bx_2 - \by_2, \bm{v} \rangle \geq 0$. Thus, the segment $[\bx_1, \bx_2]$ meets the two hyperplanes that are orthogonal to $\bm{v}$ and that go through $\by_1$ and $\by_2$, respectively.
\end{proof}

\begin{lemma}
\label{lem::projopt} Let $\mathcal{Y}$ be a non-empty closed set. For any $\bx$, we denote its projection onto $\mathcal{Y}$ by $\by^{*}$, i.e.  
    $\by^{*} = \arg \min_{\by \in \mathcal{Y}} \|\bx - \by \|_2^2$. Then for any $\by' \in \mathcal{Y}$ we have 
    $$
    \|\by^{*} - \by' \|_2^2 \leq  2\langle \by^{*} - \by',  \bx - \by' \rangle.
    $$
\end{lemma}

\begin{proof}
By the optimality of $\by^{*}$ and $\by' \in \mathcal{Y}$, we have
\begin{align*}
    \|\bx - \by^{*} \|_2^2 \leq  \|\bx - \by' \|_2^2.
\end{align*}
This implies that
\begin{align*}
    & \|\bx - \by' \|_2^2 + 2 \langle \bx - \by' , \by' - \by^{*} \rangle +   \|\by' - \by^{*} \|_2^2 \leq  \|\bx - \by' \|_2^2,
\end{align*}
which yields the desired inequality.
\end{proof}

The following lemma is a standard result from concentration inequalities \citep{boucheron2003concentration}.
\begin{lemma}
\label{lem::maxgaussian} Let $X_1, \ldots, X_n$ be $\mathcal{N}(0, \sigma^2)$ normal random variables with zero mean. Then
$
\mathbb{P}\left\{\max _{1 \leq i \leq n} X_i \geq \sqrt{2 \sigma^2(\log n+t)}\right\} \leq \exp(-t).
$
\end{lemma}
\begin{proof} 
Let $u\defined \sqrt{2 \sigma^2(\log n+t)}$. We have
\begin{align*}
    \mathbb{P}\left\{\max _{1 \leq i \leq n} X_i \geq u\right\} &=\mathbb{P}\left\{\exists i, X_i \geq u\right\} \\
    &\leq \sum_{i=1}^n \mathbb{P}\left\{X_i \geq u\right\} \\
    &\leq n \exp\left(-\frac{u^2}{2 \sigma^2}\right) = \exp(-t),
\end{align*}
which yields the desired inequality.
\end{proof}

Next, we provide the proof of Theorem ~\ref{thm::proj_uti_append}
\begin{proof}
    We define $N = n_1^{\text{syn}} \cdot n_2^{\text{syn}}$. Additionally, we define the projection set with respect to $\hat{\bQ}$ as
    \begin{align}
    \label{eq::defn_K}
        \mathcal{K} = \left\{ \frac{1}{m^{\text{syn}}}\hat{\bQ} \bb \ \Big| \ \bb \in \mathbb{R}^{N}, \bm{1}^T \bb = m^{\text{syn}},  0 \leq \bb \leq 1 \right\}.
    \end{align}
    Recall that $\bB^{*}$ is the optimal bi-adjacency matrix achievable for the synthetic individual tables without privacy constraint and $\bB^{\text{syn}}$ is the output of Algorithm~\ref{alg::adapt_alg_l2}. 
    We denote
    \begin{align*}
    \bb^* \defined \operatorname{vec}(\bB^{*}),\quad
    \hat{\bb} \defined \operatorname{vec}(\bB^{\text{syn}}).
    \end{align*}
    By definition,
    \begin{align*}
    \bb^{*} &= \argmin_{\hat{\bQ} \bb/m^{\text{syn}} \in \mathcal{K}} \left\| \frac{1}{m^\text{syn}} \hat{\bQ}\bb-\ba \right\|_2^2,\\
    \hat{\bb} &= \argmin_{\hat{\bQ} \bb/m^{\text{syn}} \in \mathcal{K}} \left\| \frac{1}{m^\text{syn}} \hat{\bQ}\bb-(\ba + \bw) \right\|_2^2,
    \end{align*}
    where $\bw$ is a Gaussian vector with zero-mean added to preserve DP guarantees, which we will specify later. Additionally, we introduce
    \begin{align}
    \label{eq::btilde}
        \tilde{\bb} &= \argmin_{\frac{1}{m^{\text{syn}}}\hat{\bQ} \bb \in\mathcal{K}} \left\|\frac{1}{m^\text{syn}} \hat{\bQ}\bb-(\frac{1}{m^{\text{syn}}}\hat{\bQ}\bb^{*} + \bw) \right\|_2^2. 
    \end{align}
    By the triangle inequality, we have:
    $$\left\| \frac{1}{m^\text{syn}} \hat{\bQ} \hat{\bb} - \ba \right\|_2 
    \leq \left\| \frac{1}{m^\text{syn}} \hat{\bQ} \hat{\bb} - \frac{1}{m^\text{syn}} \hat{\bQ} \tilde{\bb} \right\|_2 + \left\| \frac{1}{m^\text{syn}} \hat{\bQ} \tilde{\bb} - \frac{1}{m^\text{syn}} \hat{\bQ} \bb^{*} \right\|_2 + \left\| \frac{1}{m^\text{syn}} \hat{\bQ} \bb^{*} - \ba \right\|_2.$$
    Since $\frac{1}{m^\text{syn}} \hat{\bQ} \hat{\bb}$ and $\frac{1}{m^\text{syn}} \hat{\bQ} \tilde{\bb}$ are the projections of $(\ba + \bw)$ and $(\frac{1}{m^\text{syn}} \hat{\bQ}\bb^{*} + \bw)$ onto the convex set $\mathcal{K}$, respectively,
    Lemma~\ref{lem::contractive} implies that
    \begin{align*}
    \left\| \frac{1}{m^\text{syn}} \hat{\bQ} \hat{\bb} - \frac{1}{m^\text{syn}} \hat{\bQ} \tilde{\bb}  \right\|_2 
    \leq \left\| (\ba + \bw) -  (\frac{1}{m^\text{syn}} \hat{\bQ} \bb^{*} + \bw)  \right\|_2  
    = \left\| \ba -  \frac{1}{m^\text{syn}} \hat{\bQ} \bb^{*} \right\|_2.
    \end{align*}
    Hence, we have
    \begin{align}
    \label{eq::app_prof_tri}
    \left\| \frac{1}{m^\text{syn}} \hat{\bQ} \hat{\bb} - \ba \right\|_2 
    \leq 2 \left\| \ba -  \frac{1}{m^\text{syn}} \hat{\bQ} \bb^{*} \right\|_2 + \left\| \frac{1}{m^\text{syn}} \hat{\bQ} \tilde{\bb} - \frac{1}{m^\text{syn}} \hat{\bQ} \bb^{*} \right\|_2.
    \end{align}
    By the definition of $\tilde{\bb}$ in \eqref{eq::btilde} and Lemma \ref{lem::projopt}, we have
    \begin{align*}
    \left\| \frac{1}{m^\text{syn}} \hat{\bQ} \tilde{\bb} - \frac{1}{m^\text{syn}} \hat{\bQ} \bb^{*} \right\|_2^2 
    &\leq \frac{2}{m^\text{syn}} \left\langle  \hat{\bQ} \tilde{\bb} - \hat{\bQ} \bb^{*}, (\frac{1}{m^\text{syn}} \hat{\bQ} \bb^{*} + \bw) - \frac{1}{m^\text{syn}} \hat{\bQ} \bb^{*} \right\rangle \\
    &= \frac{2}{m^\text{syn}} \langle \hat{\bQ} (\tilde{\bb} - \bb^{*}), \bw \rangle.
    \end{align*}
    Since $\frac{1}{m^{\text{syn}}}\hat{\bQ}\bb^{*}, \frac{1}{m^{\text{syn}}}\hat{\bQ}\tilde{\bb} \in \mathcal{K}$, we have $ \| \bb^{*} \|_1 = \| \tilde{\bb}\|_1 = m^{\text{syn}}$. 
    We denote the $j$-th column of $\hat{\bQ}$ by $\bp_j$ for $j = 1, \cdots, N$. Then
    \begin{align*}
    \left\| \frac{1}{m^\text{syn}} \hat{\bQ} \tilde{\bb} - \frac{1}{m^\text{syn}} \hat{\bQ} \bb^{*} \right\|_2^2 
    & \leq \frac{2}{m^\text{syn}} \left\langle \hat{\bQ} (\tilde{\bb} - \bb^{*}), \bw \right\rangle \\
    & = \frac{2}{m^\text{syn}} \sum_{j = 1}^{N} (\tilde{b}_j - b^{*}_j) \cdot \left\langle \bp_j, \bw \right\rangle \\
    & \leq \frac{2}{m^\text{syn}} \sum_{j = 1}^{N} |\tilde{b}_j - b^{*}_j| \cdot \max_{j = 1}^{N} |\left\langle \bp_j, \bw \right\rangle|   \\
    & \leq \frac{2}{m^\text{syn}} \cdot (\|\tilde{\bb}\|_1 + \| \bb^{*} \|_1) \cdot \max_{j = 1}^{N} |\left\langle \bp_j, \bw \right\rangle|   \\
    & = 4 \cdot \max_{j = 1}^{N} |\left\langle \bp_j, \bw \right\rangle|.
    \end{align*}
    We introduce random variables:
    $$X_j \defined \langle \bp_j, \bw \rangle,\ X_{N+j} \defined - \langle \bp_j, \bw \rangle, \quad \text{for } j = 1, \cdots, N.$$
    Since each random variable $X_j$ for $j\in[2N]$ is a linear combination of independent zero mean Gaussian random variables, $X_j$ is also a zero mean Gaussian random variable.
    Next, we prove that $\bp_j$ has $|\mathcal{W}_{\text{cross},k}|$ elements equal to $1$ and the remaining elements being $0$. Here $|\mathcal{W}_{\text{cross},k}|$ represents the number of (cross-table) $k$-way workloads. 
    Note that there exist $r \in [n^{\text{syn}}_1], s \in [n^{\text{syn}}_2]$ such that
    \begin{align*}
        j = (r-1) n^{\text{syn}}_2 + s.
    \end{align*}
    Recall the definition of query vector in Lemma~\ref{lem::qv}. 
    The $i$-th element of $\bp_j$ is an indicator of whether the value of the $i$-th cross-table $k$-way marginal query of $(\bx^{\text{syn}}_r, \bx^{\text{syn}}_s)$ is $1$. Here $\bx^{\text{syn}}_r$ and $\bx^{\text{syn}}_s$ are the $r$-th and $s$-th records of $\mathcal{D}_1^{\text{syn}}$ and $\mathcal{D}_2^{\text{syn}}$, respectively. There are $|\mathcal{W}_{\text{cross},k}|$ workloads in total. Within each workload, for queries associated with it, only one query can have a value of $1$ for $(\bx^{\text{syn}}_r, \bx^{\text{syn}}_s)$. Hence, there are $|\mathcal{W}_{\text{cross},k}|$ elements of $\bp_j$ being $1$.
    Then for $j = 1,\cdots,N$,
    \begin{align*}
    \Var{X_{N+j}} &= \Var{X_j}\\
    &= \Var{\langle \bp_j, \bw \rangle} = |\mathcal{W}_{\text{cross}, k}| \cdot \Var{w_i}.
    \end{align*}
    Lemma~\ref{lem::maxgaussian} implies that with probability at least $1 - \beta$, we have
    \begin{align*}
        \left\|\frac{1}{m^\text{syn}} \hat{\bQ}\tilde{\bb}- \frac{1}{m^\text{syn}} \hat{\bQ}\bb^{*} \right\|_2^2
        &\leq 
        4 \cdot \max_{j = 1}^{2N} X_j \\
        &\leq  4 \cdot \sqrt{2 |\mathcal{W}_{\text{cross}, k}|\cdot \Var{w_i} \left(\operatorname{log} (2 N) +\operatorname{log}\frac{1}{\beta}\right)}.
    \end{align*}
    Substituting the above inequality into \eqref{eq::app_prof_tri} yields
    \begin{align*}
        \left\| \frac{1}{m^\text{syn}} \hat{\bQ} \hat{\bb} - \ba \right\|_2 
        \leq 2\left\| \ba -  \frac{1}{m^\text{syn}} \hat{\bQ} \bb^{*} \right\|_2 + \left(32 |\mathcal{W}_{\text{cross}, k}| \cdot \Var{w_i} \left(\operatorname{log} (2 N) +\operatorname{log}\frac{1}{\beta}\right)\right)^{\frac{1}{4}}.
    \end{align*}
    Since we run Algorithm~\ref{alg::adapt_alg_l2} (without unbiased sampling) for a single iteration across all $k$-way workloads, we have $T = 1$, $K = |\mathcal{W}_{\text{cross}, k}|$, and $\alpha = 0$. Therefore,
    $$\Var{w_i}
    = \left(\frac{\sqrt{2} d_{\text{max}}}{m \epsilon_0} \right)^2
    = K \cdot \frac{d_{\text{max}}^2}{m^2 \rho_{\text{rel}}}
    = |\mathcal{W}_{\text{cross}, k}| \cdot \frac{d_{\text{max}}^2}{m^2 \rho_{\text{rel}}}.$$
    As a result,
    \begin{align*}
        \left\| \frac{1}{m^\text{syn}} 
        \hat{\bQ} \hat{\bb} - \ba \right\|_2^2
        \leq 8 \left\| \frac{1}{m^\text{syn}}\hat{\bQ}\bb^{*} - \ba\right\|_2^2 + 8\sqrt{2}|\mathcal{W}_{\text{cross}, k}|\frac{d_{\text{max}}}{m\sqrt{\rho_{\text{rel}}}} \left( \operatorname{log}\frac{2 N}{\beta}\right)^{\frac{1}{2}}.
    \end{align*}
\end{proof}
